\documentclass[10pt,twocolumn,letterpaper]{article}

\usepackage{cvpr}              % To produce the CAMERA-READY version
\definecolor{cvprblue}{rgb}{0.21,0.49,0.74}
\usepackage[pagebackref,breaklinks,colorlinks,allcolors=cvprblue]{hyperref}
\newcommand{\Tref}[1]{Table~\ref{#1}}

\newcommand{\fref}[1]{Fig.~\ref{#1}}

\usepackage{xcolor}
\newcounter{todos}
\AtEndDocument{\ifnum\value{todos}>0 \PackageWarning{TODOS}{There are \arabic{todos} todos left in this paper! Fix them before submitting the paper!} \fi}

\newcommand{\V}[1]{\ensuremath{\mathbf{#1}}}

\makeatletter
\DeclareRobustCommand\onedot{\futurelet\@let@token\@onedot}
\def\@onedot{\ifx\@let@token.\else.\null\fi\xspace}
\def\eg{\emph{e.g}\onedot} 
\def\ie{\emph{i.e}\onedot}

\makeatother
\usepackage[utf8]{inputenc} % allow utf-8 input
\usepackage[T1]{fontenc}    % use 8-bit T1 fonts
\usepackage{url}            % simple URL typesetting
\usepackage{booktabs}       % professional-quality tables
\usepackage{amsfonts}       % blackboard math symbols
\usepackage{nicefrac}       % compact symbols for 1/2, etc.
\usepackage{xcolor}         % colors
\usepackage{graphicx}
\usepackage{amsmath}
\usepackage{amssymb}
\usepackage{bm}
\usepackage{makecell}
\usepackage{multirow}
\usepackage{xspace}
\usepackage{colortbl}
\usepackage{wrapfig}
\usepackage{lipsum} % Just for generating dummy text, can be removed
\usepackage{threeparttable}

\definecolor{MyDarkRed}{rgb}{0.46, 0.16, 0.16}
\definecolor{MyDarkBlue}{rgb}{0.16, 0.16, 0.66}
\definecolor{MyPink}{rgb}{1.0, 0.702, 0.729} % #ffb3ba (255,179,186)
\definecolor{MyPeach}{rgb}{1.0, 0.875, 0.729} % #ffdfba (255,223,186)
\definecolor{MyLightYellow}{rgb}{1.0, 1.0, 0.729} % #ffffba (255,255,186)
\definecolor{MyLightGreen}{rgb}{0.729, 1.0, 0.788} % #baffc9 (186,255,201)
\definecolor{MyLightBlue}{rgb}{0.729, 0.882, 1.0} % #bae1ff (186,225,255)

\newcommand{\correct}{\textcolor{MyDarkBlue}{\checkmark \xspace}}
\newcommand{\wrong}{\textcolor{MyDarkRed}{$\times$ \xspace}}

\newcommand{\dmv}{DiLiGenT-MV~\cite{bmvps}\xspace}
\newcommand{\OwL}{Objects with Lighting~\cite{old}\xspace}
\newcommand{\ReNe}{ReLight My NeRF~\cite{rene}\xspace}
\newcommand{\openillumination}{OpenIllumination~\cite{openillumination}\xspace}
\newcommand{\ORB}{Stanford-ORB~\cite{stanfordorb}\xspace}

\newcommand{\mvimgnet}{MVImageNet~\cite{mvimgnet}\xspace}
\newcommand{\mobilebrick}{Mobilebrick~\cite{mobilebrick}\xspace}
\newcommand{\cod}{CO3D~\cite{co3d}\xspace}
\newcommand{\DTUMVS}{DTU~\cite{dtumvs}\xspace}
\newcommand{\ours}{EvalMVX\xspace}

\newcommand{\SkoltechDDD}{Skoltech3D\cite{voynov2023multi}\xspace}
\newcommand{\OmniObjectDDD}{OmniObject3D\cite{wu2023omniobject3d}\xspace}

\newcommand{\neus}{NeuS~\cite{neus}\xspace}
\newcommand{\colmap}{COLMAP~\cite{colmap}\xspace}
\newcommand{\pandora}{PANDORA~\cite{pandora}\xspace}
\newcommand{\nersp}{NeRSP~\cite{nersp}}
\newcommand{\neisf}{NeISF~\cite{neisf}\xspace}
\newcommand{\mvas}{MVAS~\cite{mvas}\xspace}

\newcommand{\psnerf}{PS-NeRF~\cite{psnerf}\xspace}
\newcommand{\rnbneus}{RnbNeuS~\cite{rnbneus}\xspace}
\newcommand{\supernormal}{SuperNormal~\cite{supernormal}\xspace}
\newcommand{\mvpsnet}{MVPSNet~\cite{mvpsnet}\xspace}

\newcommand{\vminer}{VMINer~\cite{vminer}\xspace}
\newcommand{\wildlight}{WildLight~\cite{wildlight}\xspace}

\newcommand{\nero}{NeRO~\cite{nero}\xspace}

\newcommand{\relightablegs}{R-3DGS~\cite{relightable3d}\xspace}
\newcommand{\gaussianS}{GaussianSurfels~\cite{gaussianS}\xspace}
\newcommand{\pisr}{PISR~\cite{pisr}\xspace}
\newcommand{\neuss}{NeuS2~\cite{neus2}\xspace}
\newcommand{\neuralangelo}{Neuralangelo~\cite{neuralangelo}\xspace}
\newcommand{\petneus}{PET-NeuS~\cite{petneus}\xspace}

\newcommand{\refneus}{Ref-NeuS~\cite{refneus}\xspace}
\newcommand{\volsdf}{VolSDF~\cite{volsdf}\xspace}
\newcommand{\svolsdf}{S-VolSDF~\cite{svolsdf}\xspace}
\newcommand{\sparseneus}{SparseNeuS~\cite{sparseneus}\xspace}
\newcommand{\pytorchddd}{Pytorch3D~\cite{ravi2020pytorch3d}\xspace}

\newcommand{\unips}{Uni-MS-PS~\cite{hardy2024uni}\xspace}

\newcommand{\uamvps}{UA-MVPS~\cite{uamvps}\xspace}
\newcommand{\neudf}{NeDUF~\cite{liu2023neudf}\xspace}

\def\paperID{4295} % *** Enter the Paper ID here
\def\confName{CVPR}
\def\confYear{2026}

\title{EvalMVX: A Unified Benchmarking for Neural 3D Reconstruction under Diverse Multiview Setups}

\author{
\begin{tabular}{c}
\textbf{Zaiyan Yang}$^{1}$ \quad \textbf{Jieji Ren}$^{2}$ \quad \textbf{Xiangyi Wang}$^{1}$ \quad \textbf{Zonglin Li}$^{1}$ \\
\textbf{Xu Cao}$^{3}$ \quad \textbf{Heng Guo}$^{1,4*}$ \quad \textbf{Zhanyu Ma}$^{1}$ \quad \textbf{Boxin Shi}$^{5}$ \\
\\
$^{1}$Beijing University of Posts and Telecommunications, Beijing, China \\
$^{2}$School of Mechanical Engineering, Shanghai Jiao Tong University, Shanghai, China \\
$^{3}$Independent Researcher \\
$^{4}$Xiong'an Aerospace Information Research Institute, Xiong'an, China \\
$^{5}$State Key Laboratory of Multimedia Information Processing, School of Computer Science \\
\qquad Peking University, Beijing, China \\
\end{tabular}
}

\begin{document}

\maketitle
\begin{abstract}
	Recent advancements in neural surface reconstruction have significantly enhanced 3D reconstruction. However, current real world datasets mainly focus on benchmarking multiview stereo~(MVS) based on RGB inputs. Multiview photometric stereo (MVPS) and multiview shape from polarization (MVSfP), though indispensable on high-fidelity surface reconstruction and sparse inputs, have not been quantitatively assessed together with MVS. To determine the working range of different MVX~(MVS, MVSfP, and MVPS) techniques, we propose EvalMVX, a real-world dataset containing $25$ objects, each captured with a polarized camera under $20$ varying views and $17$ light conditions including OLAT and natural illumination, leading to $8,500$ images. 
	Each object includes aligned ground-truth 3D mesh, facilitating quantitative benchmarking of MVX methods simultaneously. Based on our EvalMVX, we evaluate $13$ MVX methods published in recent years, record the best-performing methods, and identify open problems under diverse geometric details and reflectance types. We hope EvalMVX and the benchmarking results can inspire future research on multiview 3D reconstruction. 
    
\end{abstract}
\section{Introduction}
\label{sec:intro}

3D reconstruction from multiview images is a foundational problem in computer vision, critical for applications such as virtual reality, appearance editing, and more. The most common setup of multiview 3D reconstruction is multiview stereo~(MVS)~\cite{neus,neus2, nero}, which takes multiview RGB images as input for recovering surface shape. Considering the complex reflectance, multiview shape from polarization~(MVSfP)~\cite{pandora, nersp, neisf} taking multiview polarization images as input is proposed, which is shown to be effective in reflective surface recovery even under sparse capture due to additional photometric and geometric cues from polarization images. Besides, multiview photometric stereo~(MVPS)~\cite{supernormal, rnbneus, uamvps} excels in detailed surface reconstruction for general reflectance, leveraging surface normal maps estimated from images under varying illumination.

With the rapid development of MVX~(\ie, MVS, MVPS, and MVSfP) techniques since the proposition of NeRF~\cite{nerf}, understanding the working range and trade-offs of each approach is essential. For instance, while both {\petneus} and {\supernormal} claim high-fidelity surface reconstruction, they differ in setup—MVS and MVPS, respectively. Identifying the balance between capture effort and reconstruction quality across setups is crucial for guiding users in selecting the optimal method and configuration for 3D surface reconstruction tasks.
Achieving this goal necessitates a dedicated design of a real-world datasets which can jointly evaluating different setups in a quantitative manner.

However, existing real-world datasets primarily benchmark only one of the MVX techniques. For example, datasets such as \DTUMVS, \mvimgnet, and \mobilebrick focus on evaluating MVS; \dmv, \openillumination, and \ReNe contain multiview and multilight images for MVPS evaluation; \nersp and \pandora capture multiview polarization images for assessing MVSfP. To the best of our knowledge, there is no dataset capable of benchmarking MVS, MVSfP, and MVPS simultaneously for identifying their application scenarios. Additionally, existing datasets~\cite{old,mvimgnet,openillumination,rene} lack aligned ground-truth (GT) shapes essential for quantitative surface reconstruction evaluation. One solution is using synthetic datasets such as NeRF-synthetic~\cite{nerf}. However, synthetic datasets struggle to replicate complex physical characteristics and noise patterns, resulting in a domain gap between synthetic evaluations and real-world performance.

\begin{figure*}
	\includegraphics[width=\linewidth]{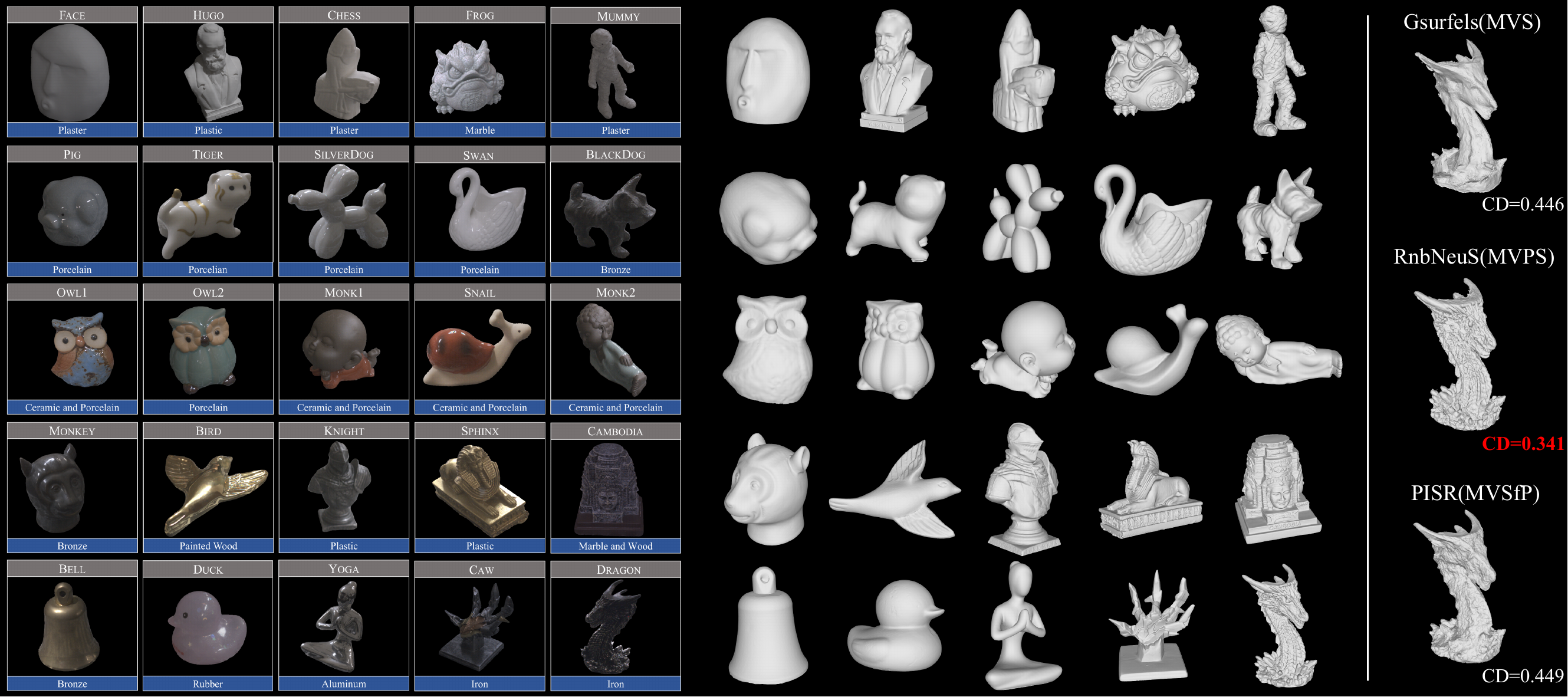}
	\caption{
    Dataset overview. \ours contains 25 objects~(left), covering diverse shapes and materials. From top to bottom, the materials range from simple diffuse to complex metallic and reflective ones, while the complexity of object shapes increases from left to right. With aligned GT shapes~(middle), we can quantitatively evaluate MVX methods on the same object~(right) and reveal their performance on detailed shape recovery and robustness against complex material.
    }
	\label{fig:teaser}
\end{figure*}

To address these problems, we present EvalMVX, the first real-world dataset with aligned GT shapes designed for quantitatively benchmarking MVS, MVSfP, and MVPS simultaneously. As shown in \fref{fig:teaser}, \ours captures 25 objects with diverse shapes and reflectances. From left to right, more details occur on surface shape, while the reflectance changes from simple diffuse to complex specular and metallic from top to bottom. For each object, we capture $20$ views using a polarization camera. Each with one image taken under environment-light and 16 images taken in a one-light-at-a-time (OLAT) manner. We scan a 3D mesh and propose an inverse silhouette rendering pipeline to align the 3D scan with the camera views.
Given multiview polarized images captured under environment light and the corresponding GT shape, we evaluate the performance of MVSfP. Additionally, using unpolarized images decomposed from the polarization data, we benchmark MVS under static environment light and MVPS under varying illuminations. In this way, all MVX methods can be quantitatively evaluated alongside \ours on the same objects observed from identical viewpoints.

Based on \ours, we benchmark $4$ MVS methods, $4$ MVPS methods, and $3$ MVSfP methods and quantitatively evaluate each in terms of shape reconstruction accuracy and computational efficiency. For each object, we also report the best-performing method to highlight the trade-offs between shape and reflectance complexity, reconstruction accuracy, and capture effort.

\vspace{- 0.2cm}

\paragraph{Contributions} This paper introduces EvalMVX, the first real-world dataset designed to evaluate MVS, MVSfP, and MVPS methods simultaneously by capturing multiview polarization images under controlled illumination. With aligned GT shapes, our benchmark provides a quantitative assessment of MVX methods, establishing their working ranges for various shapes and reflectances.

\section{Related works}
\label{sec:related_work}

We briefly summarize recent advancements in multiview stereo (MVS), multiview photometric stereo (MVPS), and multiview shape from polarization (MVSfP), highlighting their input settings and focuses. Additionally, we review real-world datasets collected for evaluating MVX methods.

\subsection{MVX methods}

\paragraph{Multiview Stereo} recovers 3D shapes from RGB images captured at different camera poses. Traditional methods like \colmap rely on Structure from Motion, requiring feature correspondences across views. Recent methods, such as \neus and \volsdf, improve MVS by leveraging differentiable volume rendering and neural implicit surface representations. For reflective surfaces, \refneus and \nero use integrated positional encoding~\cite{refnerf} to disentangle shape, reflectance, and illumination. To handle sparse views, \sparseneus and \svolsdf incorporate generalizable priors from image features and probability volume to gain the robustness of estimated shape. \petneus and \neuralangelo apply positional encoding tri-planes and a coarse-to-fine optimization strategy on hash grids, shown to be effective in recovering geometric details.  
For efficiency, \neuss integrates multi-resolution hash encoding and a CUDA implementation to accelerate \neus. Meanwhile, \relightablegs and \gaussianS accelerate reconstruction by incorporating 3D Gaussian Splatting (3DGS)~\cite{3Dgaussians}.

\vspace{- 0.2cm}

\paragraph{Multiview photometric stereo} recovers surface shape from images captured under varying light and viewpoints, offering advantages for detailed shape recovery. \psnerf estimates surface normals from multi-light images using photometric stereo, then uses them to regularize the neural radiance field.  
\uamvps address unreliable surface estimates problem by introducing uncertainty. \supernormal and \rnbneus first estimate per-view surface normals with off-the-shelf photometric stereo method~\cite{sdm} and refine 3D surface shape through mutliview surface normal consistency. Instead of per-view multi-light capture, \vminer and \wildlight capture flash and no-flash image pairs at varying viewpoints, improving shape reconstruction accuracy while maintaining a practical setup. Unlike above methods that require per-scene optimization, \mvpsnet trains a generalizable MVPS network capable of efficient depth and normal prediction.

\vspace{- 0.2cm}

\paragraph{Multiview shape from polarization} offers advantages over MVS as angle of polarization~(AoP) provides geometric cue of surface normals. Based on this property, \pandora achieves shape recovery and radiance decomposition by minimizing the difference between rendered and observed Stokes vectors. \mvas leverages azimuth angle consistency across views to recover surface shape from AoP. \nersp combines the photometric cue and geometric cue from polarization images and achieves reflective surface reconstruction under sparse polarization. \pisr boosts the efficiency of MVSfP by introducing hash-grid-based neural SDF. \neisf proposes a neural implicit Stokes field and shows the merit of MVSfP in handling scenes containing strong inter-reflections.

\subsection{MVX datasets}

\begin{table*}
	\caption{Summary of MVX datasets and their contents.}
	\resizebox{\linewidth}{!}{
		\begin{threeparttable}
			\begin{tabular}{cccccccc}
				\toprule
				Dataset & EvalMVS & EvalMVPS & EvalMVSfP & 3D GT & \#Objects & \#Lights & \#Views \\
				\midrule
				\DTUMVS& \correct& \wrong& \wrong& Point cloud& 80& 1/7 Env.\tnote{a} & 49$\sim$64\\
                    \mobilebrick & \correct& \wrong& \wrong& Mesh& 153& 1 Env. &30$\sim$134\\
                    \ORB& \correct& \wrong& \wrong& Mesh& 14& 3/7 Env.&70\\
                    \OwL & \correct& \wrong& \wrong& None& 8& 3/72 Env. & 42$\sim$67\\
                    \cod & \correct& \wrong& \wrong& Point cloud& 19K& Env.~(Case dependent)& Case dependent\\
				\mvimgnet& \correct& \wrong& \wrong& None& 220K& Env.~(Case dependent)& Case dependent\\
                    \OmniObjectDDD & \correct& \wrong& \wrong& Mesh& 6k& Env.~(Case dependent) & 100\\

				\dmv & \wrong & \correct & \wrong & Mesh& 5& 96 OLAT\tnote{b}  & 20\\
				\openillumination & \wrong& \correct& \wrong& None& 64& 13 PAT\tnote{c} + 142 OLAT&72\\
				\ReNe & \wrong& \correct& \wrong& None& 20& 40 OLAT&50\\
                    \SkoltechDDD & \correct& \correct& \wrong& Mesh& 107& 14 Env. & 100\\
				
				PANDORA Dataset~\cite{pandora}& \correct& \wrong& \correct& None& 3& 1 Env.&35\\
				NeRSP Dataset~\cite{nersp} & \correct& \wrong& \correct& Mesh& 5& 1 Env.&6\\
				PISR Dataset~\cite{pisr}& \correct& \wrong& \correct& Mesh& 4& 1 Env.&40\\
				\rowcolor{lightgray}
				\ours & \correct & \correct & \correct & Mesh& 25 & 16 OLAT + 1 Env. & 20 \\
				\bottomrule
			\end{tabular}
			\begin{tablenotes}
				\item \parbox{\linewidth}{[a] Env: Environment light;\quad [b] OLAT: One light at a time; \quad [c] PAT: Pattern illumination.}
			\end{tablenotes}
		\end{threeparttable}
	}
	\label{tab:dataset_summary}
\end{table*}

As shown in \Tref{tab:dataset_summary}, we summarize public real-world datasets for evaluating MVX in recent years. Below, we briefly describe each dataset, categorized by its target benchmark tasks.

\vspace{- 0.2cm}

\paragraph{Multiview stereo.} \DTUMVS is commonly used to benchmark MVS methods under environment illuminations. \ORB scans $14$ objects and aligns scanned meshes with captured multiview images for quantitative evaluation. \mobilebrick captures images of $153$ LEGO models where 3D GT shapes can be obtained without using a scanner. \OwL collects images captured under varying poses and illuminations to evaluate the performance of MVS at different natural scenes. 
To expand the scale of MVS dataset, \mvimgnet captures $220K$ objects from $238$ classes, leading to $6.5$ million frames from $219,188$ videos. \cod collects $1.5$ million frames from $19,000$ videos across $50$ object categories with accurate camera poses and the GT 3D point clouds. \OmniObjectDDD provides a large vocabulary of $6,000$ high-quality real-scanned 3D objects across $190$ categories. 

\vspace{- 0.2cm}
 
\paragraph{Multiview photometric stereo.} \dmv is the first real-world dataset dedicated to benchmarking MVPS, comprising 5 objects captured from 20 views, with each view illuminated by 96 OLAT sources in a darkroom. \ReNe and \openillumination expand the scale of \dmv to $20$ and $64$ objects, respectively. However, the GT shapes are not provided for quantitative evaluation. \SkoltechDDD offers $1.4$ million images of $107$ scenes captured under $100$ views and $14$ lighting conditions using $7$ different sensors~(\eg, Kinect, structured-light scanner), covering diverse shapes and materials. The above datasets can evaluate MVPS based on general unknown illumination settings~\cite{sdm}.

\vspace{- 0.2cm}

\paragraph{Multiview shape from polarization.} Different from MVS and MVPS, benchmarking MVSfP requires specific polarization camera, making the scale of existing real-world MVSfP dataset small. \pandora captures $3$ objects, each captured with $35$ views under natural illumination with a snapshot polarization camera. However, the GT shapes are not provided. To handle this problem, \nersp and \pisr capture 4 objects each, aligning GT shapes with the captured images. By using a polarization image formation model, unpolarized images can be extracted from the captured polarization ones, allowing these datasets to support both MVSfP and MVS benchmarking.

Despite substantial efforts in real-world dataset collection, quantitatively evaluating MVS, MVPS, and MVSfP simultaneously remains an open problem. This requires aligned GT 3D shapes, varied illuminations from both environment light and OLAT, and a polarization camera. Our \ours is shown to be the first dataset to address these needs by capturing 25 objects with a snapshot polarization camera under 16 OLATs and one static environment light.

\section{\ours}
\label{sec:dataset}

This paper aims to benchmark MVX methods comprehensively across various configurations. From a sensor perspective, a polarization camera is essential for evaluating both MVSfP and MVS methods. From an illumination standpoint, images captured from the same view under varying OLAT and static environment light are necessary to assess both MVS and MVPS methods. Finally, for shape and reflectance, selecting objects that range from simple, smooth shapes with diffuse reflectance to complex, detailed shapes with specular reflectance allows us to effectively determine the working ranges of different MVX methods.

\subsection{Data capture}
\begin{figure*}
        \centering
	\includegraphics[width=\linewidth]{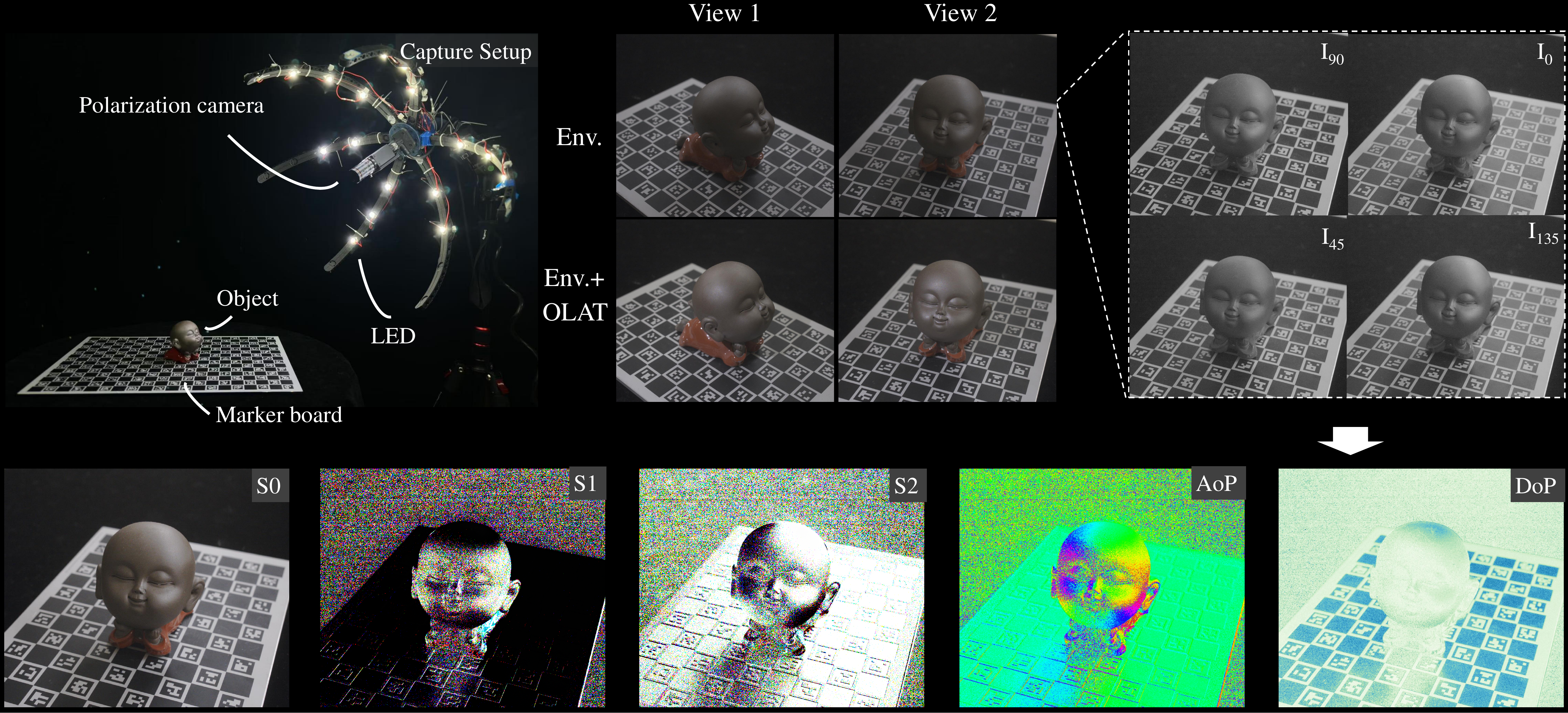}
	\caption{Capture setup and data processing of \ours. (\textbf{Top}) Our capture system contains a polarization camera and 16 evenly distributed LEDs for recording multiview and multi-light images. (\textbf{Bottom}) Polarization information can be extracted from the snapshot polarization image.}
	\label{fig:capture_setup}
\end{figure*}

\paragraph{Capture Setup.} As shown in \fref{fig:capture_setup}, the object is placed on a marker board for intrinsic and extrinsic camera calibration. A 3D-printed rig holds 16 high-power LEDs arranged to maintain fixed relative positions between the LEDs and the camera. We use a Lucid Triton RGB polarization camera\footnote{\url{https://thinklucid.com/product/triton-5-mp-polarization-camera/}} equipped with a 16mm lens to capture snapshot polarization images under different viewpoints and illumination conditions.
For each view, we first capture a polarization image at a resolution of $2448 \times 2048$ under environment light. Then, we sequentially capture images with each LED toggled on and off. By subtracting the environment-light image, we obtain OLAT images for MVPS. After the multi-illumination capture at one viewpoint, we move the light-camera rig around the object and repeat the process. The object remains static throughout this multiview, multi-light acquisition. \emph{Further details about camera poses and LEDs positions are provided in supplementary material.}

\vspace{- 0.2cm}

\paragraph{Object selection.} As shown in \fref{fig:teaser}, we choose $25$ objects with diverse reflectances, containing \textbf{diffuse} surfaces such as {\sc Face} and {\sc Mummy}; \textbf{specular} surfaces such as {\sc Pig} and {\sc Monkey}; \textbf{metallic} surfaces such as {\sc Bell}; and a \textbf{translucent} surface {\sc Duck}. The objects are roughly ordered from top to bottom by increasing reflectance complexity. 
The object shapes in \ours span a range from smooth surfaces, such as {\sc Face} and {\sc Pig}, to highly detailed, complex geometries like {\sc Cambodia} and {\sc Dragon}.
% The box plot in \fref{fig:curvature_visual} quantifies the shape complexity across different objects in \ours, using average curvature~\cite{heep2024adaptive} calculated on surface normal maps as the metric. 

\subsection{Data processing}
\paragraph{Polarized image processing.}
We use a snapshot polarization camera to capture 16-bit raw image. 
Due to the structure of the Bayer filter, the raw image records image observations under different polarized directions of light, specifically at angles of \(0^\circ\), \(45^\circ\), \(90^\circ\), and \(135^\circ\), leading to four directional polarized images after demosaicing. As shown in \fref{fig:capture_setup}, based on these four images, we calculate stokes vectors \((s_0, s_1, s_2, s_3)\), which are defined as 
\begin{eqnarray}
	\begin{cases}
		s_0 = \frac{1}{2} \left( I_0 + I_{45} + I_{90} + I_{135} \right), \\
		s_1 = I_0 - I_{90}, \\
		s_2 = I_{45} - I_{135},
	\end{cases}
\end{eqnarray}
where $s_0$ represents the total intensity of the light. The elements $s_1$ and $s_2$ describe the linear polarization components in the horizontal-vertical and $+45^\circ / -45^\circ$ directions, respectively. 
We neglect circularly polarized light and set $s_3 = 0$. Given Stokes vectors, we compute AoP $\phi$ and degree of polarization (DoP) as follows:
\begin{eqnarray}
	\phi = \frac{1}{2} \arctan\left(\frac{s_2}{s_1}\right),\quad  \rho = \frac{\sqrt{s_1^2 + s_2^2}}{s_0}.
\end{eqnarray}
The polarization images, stokes vectors, and the AoP and DoP maps are used as input for MVSfP.

\vspace{- 0.2cm}

\paragraph{Mask segmentation and camera poses calibration.}
Since existing MVX methods such as \supernormal require multiview object masks as input, we employed Segment Anything Model (SAM)~\cite{kirillov2023segment} to perform silhouette mask segmentation. We use Metashape\footnote{Agisoft (2024) Agisoft Metashape. \url{https://www.agisoft.com/}} to jointly estimate the camera's intrinsic and extrinsic parameters, specifying the physical size of an Aruco Marker calibration board to recover the actual scale of the reconstructed shape.

\subsection{GT shape alignment}
% \begin{figure*}
% 	\includegraphics[width=\linewidth]{imgs/curvature_visual.pdf}
% 	\caption{(\textbf{Left}) We align scanned mesh to camera views guided by GT masks. (\textbf{Right}) Curvature distribution of \ours calculated from GT normals quantifies the shape complexity of each object.} 
% 	\label{fig:curvature_visual}
% \end{figure*}

 % As shown in \fref{fig:curvature_visual}, 
 We first scan the surface shape using a shining 3D EinScan-SP scanner\footnote{EinScan SP Scanner~\url{https://www.einscan.com/einscan-sp/}}. Then we perform a coarse alignment between the scanned mesh and the dense reconstruction generated by Metashape using ICP algorithm~\cite{besl1992method}. To further align camera poses \([\hat{\V{R}}_i, \hat{\V{t}}_i]\) with the scanned mesh, we employ a differentiable render \pytorchddd to minimize the difference between the rendered and observed object mask, \ie
\begin{equation}
    \underset{\{\hat{\mathbf{R}}_i, \hat{\mathbf{t}}_i\}_{i=1}^{K}}{\min}
    \sum_{i=1}^{K} \mathrm{BCE}\left( f_M(\mathbf{X}; \hat{\mathbf{R}}_i, \hat{\mathbf{t}}_i), \mathbf{M}_i \right),
\end{equation}
where $BCE(\cdot)$ denotes the binary cross entropy loss to measure the difference between rendered masks by projecting mesh $\V{X}$ at different camera poses and the corresponding observed mask $\V{M}_i$. Given aligned mesh and poses, we use Blender~\cite{blender} to render surface normal maps as the ``GT''. \emph{More detail of our GT alignment process and the quantitative evaluation of the alignment  quality can be found in our \textbf{supplementary material}.}

\section{Benchmark evaluations}
\label{sec:benchmark}
This section showcases the benchmark results for MVX methods using \ours. 
% \emph{Complete qualitative evaluations and mesh visualizations can be found in our supplementary material}.

\subsection{Baseline methods \& evaluation metric}

We select \neuss, an enhanced version of classical MVS method \neus, along with its extensions for detailed surface recovery (\petneus) and reflective surface recovery (\nero). In addition, we include \gaussianS adopting 3D Gaussian Splatting (3DGS)~\cite{3Dgaussians} to speed up MVS. For MVPS, we choose the state-of-the-art (SOTA) methods \supernormal and \rnbneus, utilizing \unips to estimate per-view surface normals from images captured under varying light conditions. We also include \wildlight and \vminer, where the flash image input is assigned as the one illuminated by the LED closest to the camera.
For MVSfP, we select the SOTA methods \mvas, \nersp, and \pisr.
% The shape reconstruction accuracy is evaluated using Chamfer Distance (CD) between the GT and estimated meshes. \emph{More details on evaluation settings and results on additional baselines are provided in the supplementary material}.
\emph{Complete qualitative evaluations mesh visualizations, evaluation results on additional baselines including \neus and \neudf, are provided in the \textbf{supplementary material}. }

\subsection{Overview of benchmark results}
\begin{table*}
    \caption{Benchmark results on \ours measured by CD in millimeter over 25 objects, where the smallest and second smallest CD values are shown in bold and underlined, respectively.}
    \centering
    \resizebox{\textwidth}{!}{
    \begin{tabular}{c|cccc|cccc|ccc}
        \toprule
        \multirow{2}{*}{Method} &  \multicolumn{4}{c|}{\cellcolor{MyPink} MVS} & \multicolumn{4}{c|}{\cellcolor{MyPeach} MVPS} & \multicolumn{3}{c}{\cellcolor{MyLightBlue} MVSfP} \\
        \cmidrule{2-12}
        & \neuss & \nero & \petneus & \gaussianS & \rnbneus & \supernormal & \vminer & \wildlight & \mvas & \nersp & \pisr \\
        \midrule
        {\sc Face} & \textbf{0.234} & 0.411 & 0.303 & 0.540 & 0.614 & 0.277 & 0.400 & \underline{0.250} & 0.389 & 0.330 & 0.308 \\
        {\sc Pig} & 0.312 & 0.282 & 0.408 & 0.720 & 0.379 & \textbf{0.256} & 0.470 & \underline{0.265} & 0.446 & 0.447 & 0.286 \\
        {\sc Owl1} & 0.327 & \underline{0.306} & 0.360 & 0.636 & 0.443 & \textbf{0.288} & 0.319 & 0.342 & 0.354 & 0.371 & 0.423 \\
        {\sc Monkey} & 0.487 & 0.345 & 0.664 & 0.646 & 0.393 & \underline{0.325} & 0.530 & 0.335 & 0.365 & 0.501 & \textbf{0.311} \\
        {\sc Owl2} & 0.383 & \underline{0.321} & 0.393 & 0.639 & 0.504 & \textbf{0.300} & 0.415 & 0.377 & 0.444 & 0.405 & 0.390 \\
        {\sc Bell} & 0.321 & \underline{0.254} & 1.255 & 0.635 & 0.575 & 0.476 & 0.399 & \textbf{0.200} & 0.411 & 0.373 & 0.292 \\
        {\sc Duck} & 0.413 & \underline{0.249} & 0.717 & 0.684 & 0.739 & 0.637 & 0.548 & 0.268 & 0.490 & 0.685 & \textbf{0.245} \\
        {\sc Monk1} & 0.527 & \underline{0.401} & 0.523 & 1.074 & 0.494 & \textbf{0.399} & 0.626 & 0.502 & 0.772 & 0.802 & 0.406 \\
        {\sc Tiger} & 0.385 & 0.382 & 0.520 & 0.703 & \underline{0.347} & \textbf{0.305} & 0.498 & 0.359 & 0.559 & 0.454 & 0.429 \\
        {\sc SilverDog} & 0.336 & \underline{0.310} & 0.432 & 0.618 & 0.355 & \textbf{0.280} & 0.555 & 0.324 & 0.520 & 0.363 & 0.313 \\
        {\sc Snail} & 0.309 & \textbf{0.264} & 0.346 & 0.515 & 0.355 & 0.314 & 0.409 & \underline{0.267} & 0.486 & 0.320 & 0.344 \\
        {\sc Swan} & 1.113 & 1.156 & 1.071 & 2.176 & \underline{0.971} & \textbf{0.886} & 1.195 & 1.212 & 1.406 & 1.089 & 1.177 \\
        {\sc Yoga} & 0.452 & \textbf{0.284} & 0.738 & 0.675 & 0.549 & 0.510 & 0.426 & \underline{0.295} & 0.489 & 0.517 & 0.410 \\
        {\sc Hugo} & 0.513 & 0.490 & 0.522 & 0.657 & \underline{0.451} & \textbf{0.387} & 0.543 & 0.543 & 0.646 & 0.572 & 0.598 \\
        {\sc Chess} & \underline{0.248} & 0.262 & 0.266 & 0.507 & 0.286 & \textbf{0.240} & 0.373 & 0.282 & 0.497 & 0.353 & 0.324 \\
        {\sc BlackDog} & 0.337 & 0.393 & 0.428 & 0.597 & 0.322 & \textbf{0.262} & 0.387 & \underline{0.320} & 0.520 & 0.348 & 0.364 \\
        {\sc Monk2} & 0.391 & 0.386 & 0.332 & 0.590 & \underline{0.324} & \textbf{0.211} & 0.448 & 0.377 & 0.473 & 0.373 & 0.383 \\
        {\sc Bird} & 0.447 & 0.396 & 0.727 & 0.552 & \underline{0.319} & 0.330 & 0.781 & \textbf{0.263} & 0.591 & 0.481 & 0.474 \\
        {\sc Caw} & 0.421 & \underline{0.367} & \textbf{0.338} & 0.618 & 0.375 & 0.383 & 0.818 & 0.369 & 0.799 & 0.652 & 0.387 \\
        {\sc Knight} & 0.394 & 0.405 & 0.795 & 0.503 & \underline{0.363} & \textbf{0.326} & 0.534 & 0.408 & 0.566 & 0.469 & 0.462 \\
        {\sc Frog} & 0.270 & 0.339 & 0.289 & 0.385 & \underline{0.238} & \textbf{0.171} & 0.395 & 0.308 & 0.483 & 0.351 & 0.352 \\
        {\sc Sphinx} & 0.525 & 0.546 & 0.517 & 0.699 & \underline{0.355} & \textbf{0.328} & 0.901 & 0.605 & 0.756 & 0.717 & 0.605 \\
        {\sc Cambodia} & 0.460 & 0.522 & 0.406 & 0.545 & \underline{0.291} & \textbf{0.287} & 0.726 & 0.578 & 0.822 & 0.637 & 0.512 \\
        {\sc Mummy} & 0.286 & 0.304 & 0.332 & 0.416 & \underline{0.172} & \textbf{0.162} & 0.269 & 0.265 & 0.436 & 0.296 & 0.283 \\
        {\sc Dragon} & 0.458 & 0.468 & 0.666 & 0.446 & \textbf{0.341} & \underline{0.387} & 0.593 & 0.479 & 0.578 & 0.511 & 0.449 \\
        Average & \cellcolor{gray!20} 0.414 & \cellcolor{gray!20} 0.394 & \cellcolor{gray!20} 0.534 & \cellcolor{gray!20} 0.671 & \cellcolor{gray!20} 0.422 & \cellcolor{gray!20} \textbf{0.349} & \cellcolor{gray!20} 0.542 & \cellcolor{gray!20} \underline{0.392} & \cellcolor{gray!20} 0.572 & \cellcolor{gray!20} 0.497 & \cellcolor{gray!20} 0.421 \\
        Median & \cellcolor{gray!20} 0.391 & \cellcolor{gray!20} 0.367 & \cellcolor{gray!20} 0.432 & \cellcolor{gray!20} 0.618 & \cellcolor{gray!20} 0.363 & \cellcolor{gray!20} \textbf{0.314} & \cellcolor{gray!20} 0.498 & \cellcolor{gray!20} \underline{0.335} & \cellcolor{gray!20} 0.497 & \cellcolor{gray!20} 0.454 & \cellcolor{gray!20} 0.387 \\
        \bottomrule
    \end{tabular}
    }  % 结束 resizebox
    \label{tab:benchmark_cd}
\end{table*}

% 主要分析Normal和AoP误差对算法可能造成的影响
% \begin{figure}
% 	\includegraphics[width=\linewidth]{imgs/normal_aop.pdf}
% 	\caption{\textbf{(Left)} Inaccurate surface normal inputs from \unips due to complex reflectance influence the shape recovery from \supernormal. \textbf{(Right)} The difference between the real-captured AoP map and azimuth map due to sensor noise degrades the performance of MVSfP methods.}
% 	\label{fig:normal_aop}
% \end{figure}
\begin{figure}
	\includegraphics[width=\linewidth]{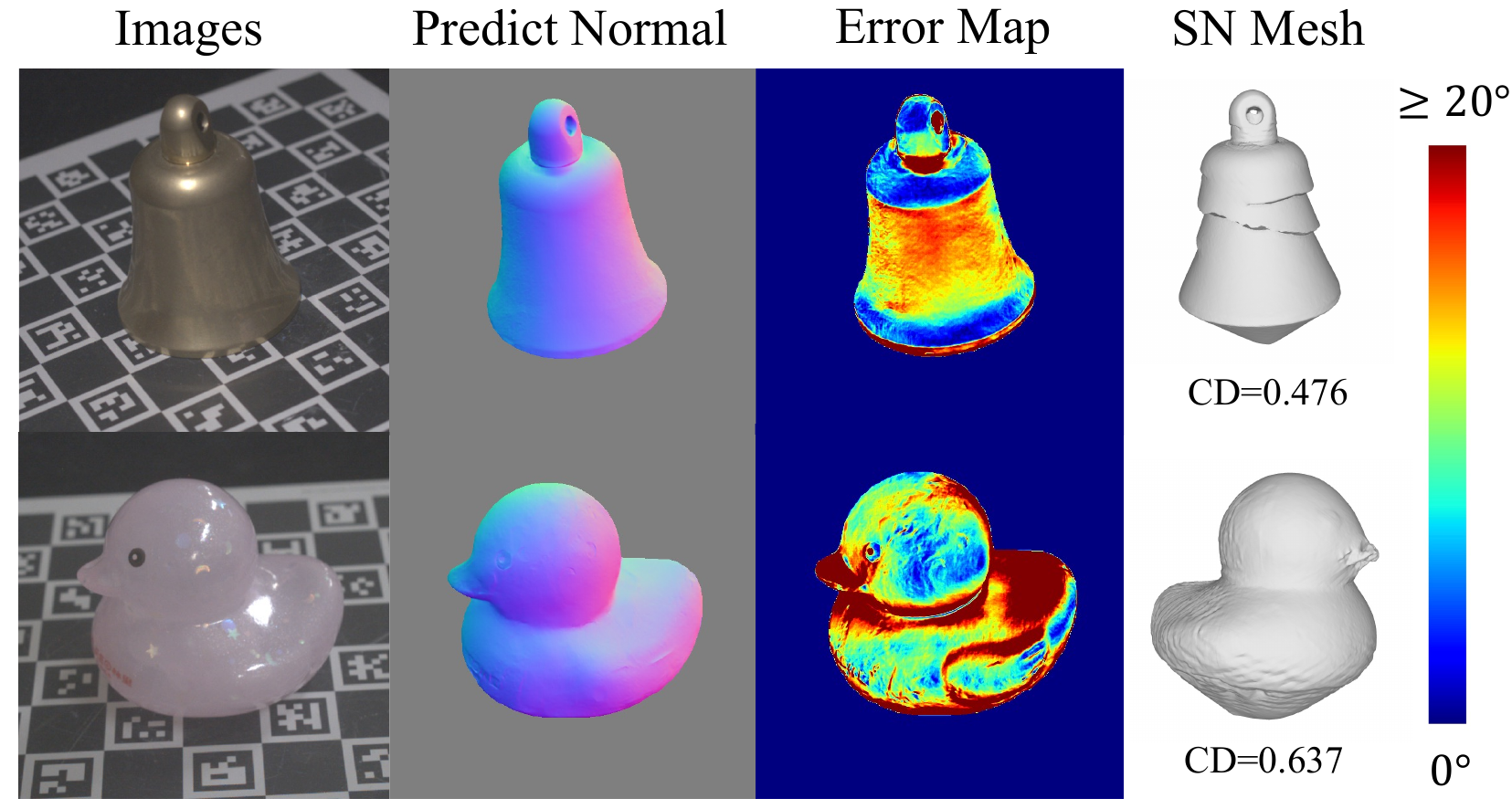}
	\caption{Inaccurate surface normal inputs from \unips due to complex reflectance influence the shape recovery from \supernormal.}
	\label{fig:normal_aop_1}
\end{figure}

\begin{figure}
	\includegraphics[width=\linewidth]{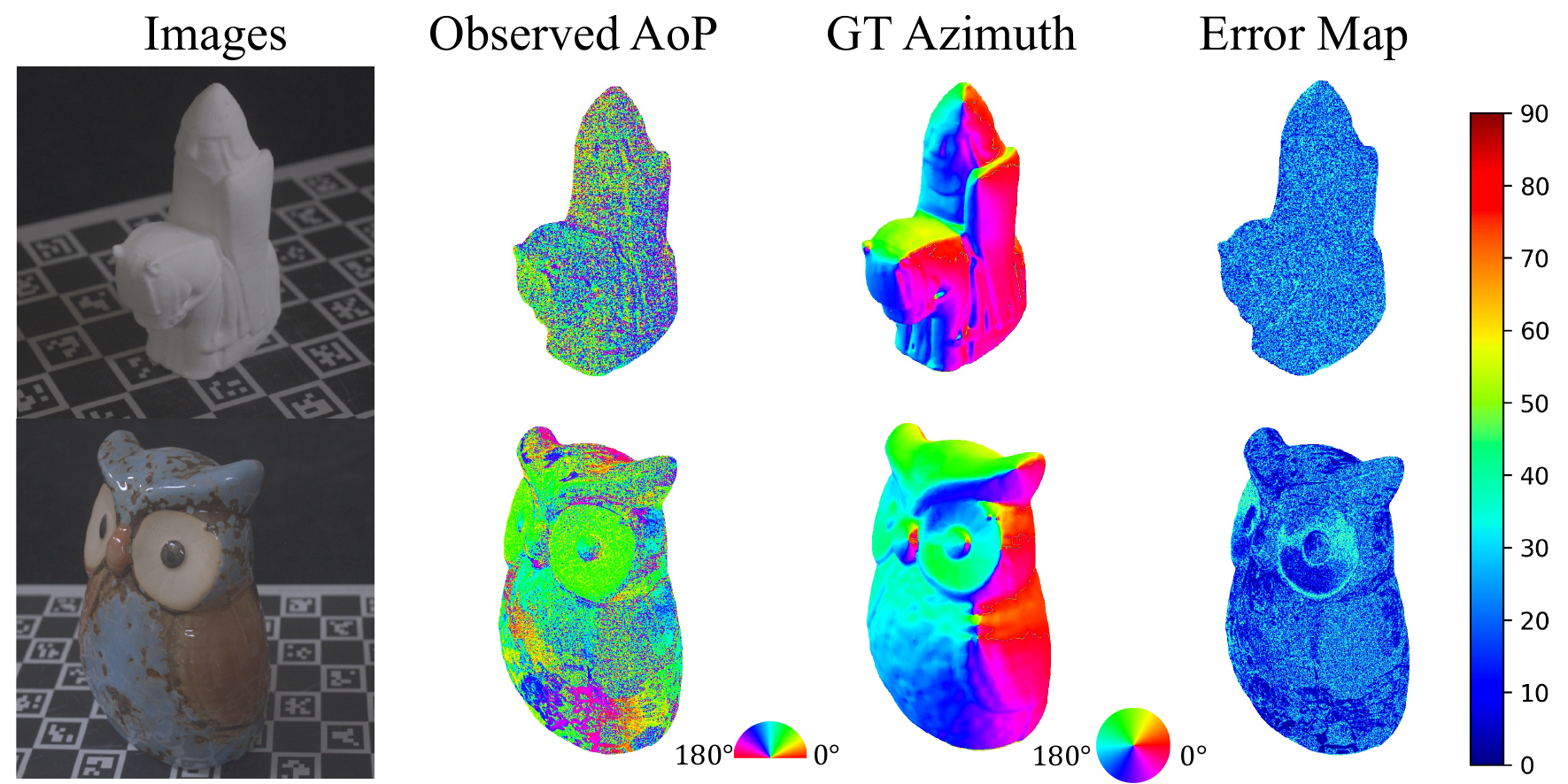}
	\caption{The difference between the real-captured AoP map and azimuth map due to sensor noise degrades the performance of MVSfP methods.}
	\label{fig:normal_aop_2}
\end{figure}

In this section, we discuss shape reconstruction results for methods at each MVX setup, followed by a working range analysis across different MVX setups. 

\vspace{- 0.2cm}

\paragraph{MVPS methods.} As shown in \Tref{tab:benchmark_cd}, \supernormal achieves the lowest or second-lowest CD on most objects, ranking as the SOTA in both median and average CD. Leveraging images captured under varying illumination, photometric stereo proves effective in handling diverse reflectance properties, and surface normals offer rich geometric cues for accurate and robust reconstruction. \supernormal integrates surface normal inputs and multi-resolution hash encoding, enabling high-fidelity surface recovery and demonstrating strong robustness to both diffuse (\eg\ {\sc Frog}) and specular (\eg\ {\sc Sphinx}) materials.

Despite the success of \supernormal, we observe failure cases on metallic objects~(\eg\ {\sc Bell}) and translucent objects~(\eg\ {\sc Duck}). As shown in \fref{fig:normal_aop_1}, the input surface normals estimated by the learning-based photometric stereo method \unips are less accurate due to complex reflectance.
This limitation also propagates to \rnbneus, which relies on \unips for surface normal input. 

On the other hand, \wildlight achieves the smallest average CD and second-smallest median CD among all methods, showing competitive results using only flash/no-flash image pairs. Unlike \supernormal, which requires multiple illuminations, \wildlight demonstrates that a minimal two-shot setup with controlled light can reduce shape-reflectance ambiguity, providing high-quality reconstruction in a practical setting. In addition, since \wildlight is a self-supervised method based on inverse rendering, the shape estimation of {\sc Bell} and {\sc Duck} is free from the error brought by the learning-based photometric stereo~\cite{hardy2024uni}. 

\paragraph{MVS Methods.} As shown in \Tref{tab:benchmark_cd}, \nero achieves the lowest or second-lowest CD on 9 out of 25 objects, particularly excelling on highly reflective metallic surfaces such as {\sc SilverDog} and {\sc Yoga}. Its use of split-sum approximation and integrated directional encoding (IDE) enables more accurate modeling of specular reflections. However, \nero performs less effectively on diffuse surfaces like {\sc Face} and {\sc Chess}, likely due to its sensitivity to static environmental illumination, as noted in the original paper. 

\petneus enhances \neus for high-fidelity surface reconstruction through positional encoding tri-planes. However, we observe that \petneus can produce overly rough meshes (\eg\ {\sc Duck})~(see supplementary material), as its highly expressive network structure may overfit and hallucinate details. Additionally, the positional encoding tri-planes module seems to make it less robust on reflective surfaces, as observed in {\sc Bird} and {\sc Bell}.
\gaussianS prioritizes fast reconstruction but produces less accurate results compared to \neuss and \nero. A key limitation is its surfel-based representation, which lacks a unified modeling of volume density, depth, and surface normals offered by SDF-based approaches.

\vspace{- 0.2cm}

\paragraph{MVSfP Methods.} Among MVSfP methods, \pisr stands out as the SOTA. Leveraging polarization images, MVSfP methods exhibit robust performance on surfaces with complex reflectance properties, such as the {\sc Monkey} and {\sc Duck}. In contrast to RGB-based methods, which often struggle with multiview consistency on non-Lambertian surfaces, AoP maps used in MVSfP inherently constrain the azimuth component of surface normals, thereby enhancing the robustness and accuracy of surface reconstruction under challenging reflectances.

However, as shown in \fref{fig:normal_aop_2}, the relationship between the observed AoP map and the azimuth map from GT surface normal does not always align with the expected differences of $\pi$ or $\pi / 2$, depending on surface reflectance and sensor noise. This inconsistency leads to noisy polarization cues, which can result in inaccurate surface reconstructions when adopted by MVSfP methods.

\vspace{- 0.2cm}

\paragraph{Method selection based on \ours.}

\begin{figure*}
	\includegraphics[width=\linewidth]{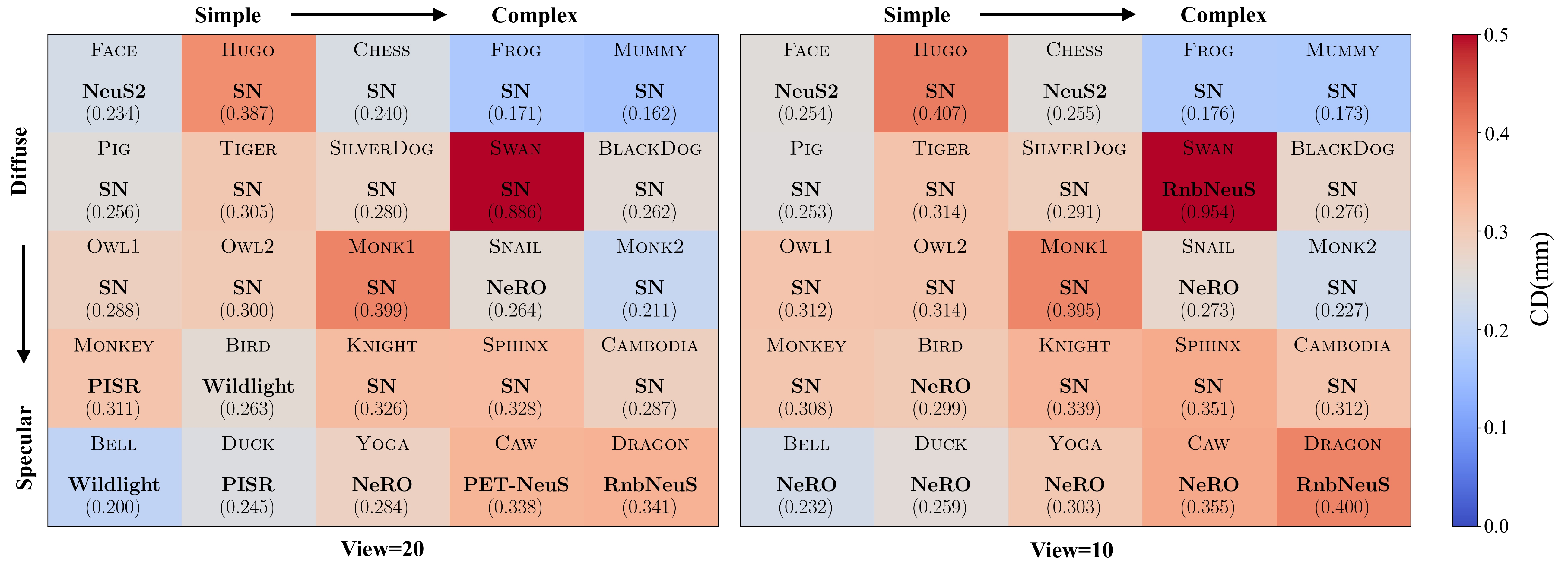}
\caption{Best-performing method for each object in
\ours, with CD displayed in brackets. 
Summary results shown in left and right correspond to using 20 and 10 input views.
}
	\label{fig:benchmark_res2}
\end{figure*}

As shown in \fref{fig:benchmark_res2}, we summarize the best-performing methods and their corresponding lowest CD values for each object in \ours under varying numbers of views, highlighting the working ranges and performance boundaries of current MVX methods.
\supernormal demonstrates a broad applicability, effectively handling a wide spectrum of surface materials—from simple diffuse to highly reflective, owing to the strengths of photometric stereo. With per-view surface normal input, \supernormal maintains top-tier accuracy even under reduced views, showing robustness to sparse observations. Moreover, since surface normals capture fine-grained geometry, MVPS methods tend to achieve the lowest CD on objects with complex shapes, as shown in the last column.
In contrast, MVSfP methods are more effective on objects with complex reflectance but relatively simple geometry, such as {\sc Duck}. MVS methods should be selected depending on surface reflectance. However, despite MVX methods get good accuracy on most object, they are still challenging for handling cast shadow~(\eg\ {\sc Swan}).

\vspace{- 0.2cm}

\paragraph{Comparison on efficiency and accuracy.}

% \begin{figure}
% 	\includegraphics[width=\linewidth]{imgs/time_acc.pdf}
% 	\caption{\textbf{(Left)} Visualization of accuracy and efficiency across different methods. \textbf{(Right)} \supernormal with integrated depth improves reconstructions considering per-view shape continuity.}
% 	\label{fig:time_acc_memory_eval}
% \end{figure}
\begin{figure}
	\includegraphics[width=\linewidth]{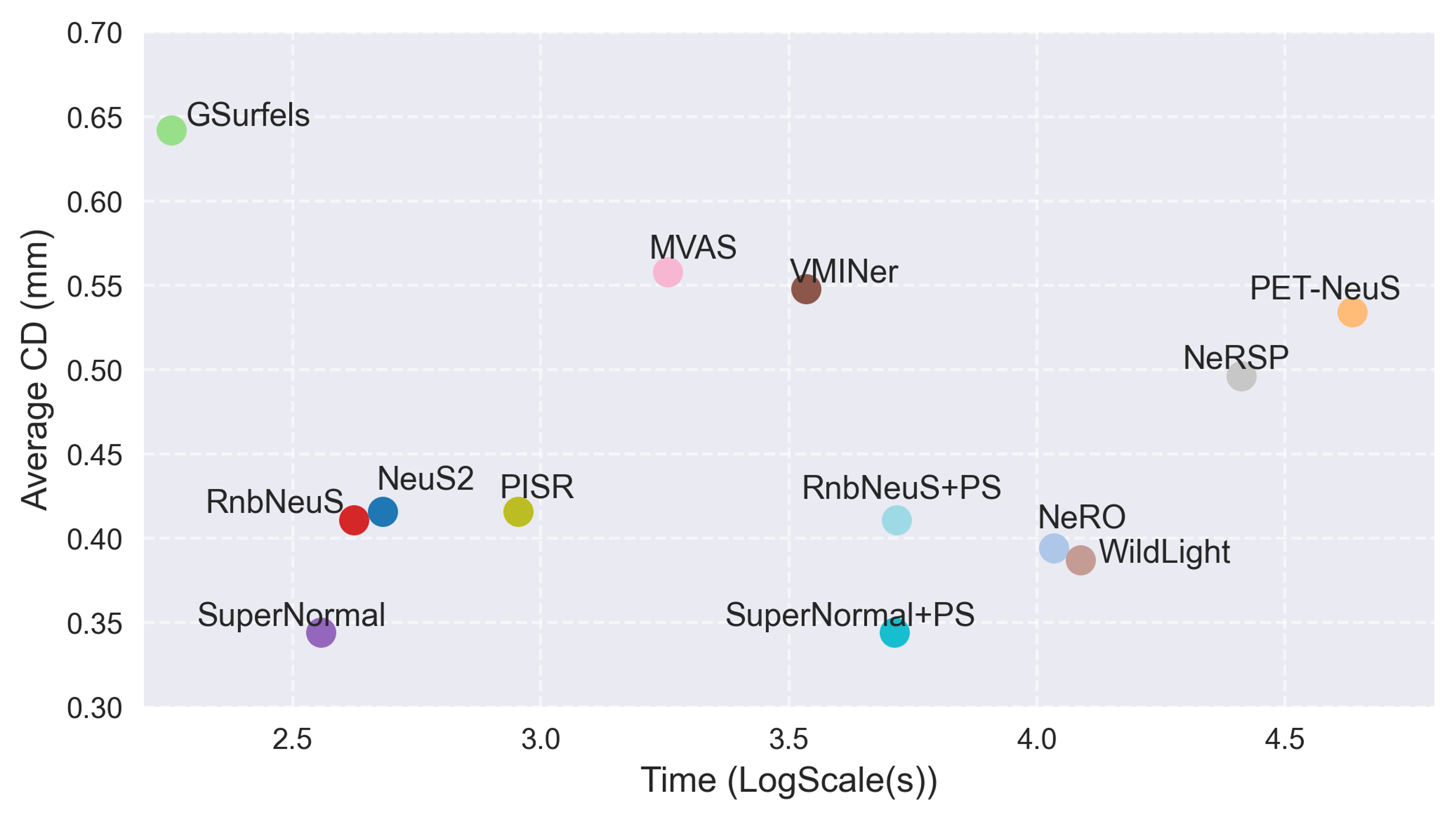}
	\caption{Visualization of accuracy and efficiency across different methods.}
	\label{fig:time_eval}
\end{figure}

\begin{figure}
	\includegraphics[width=\linewidth]{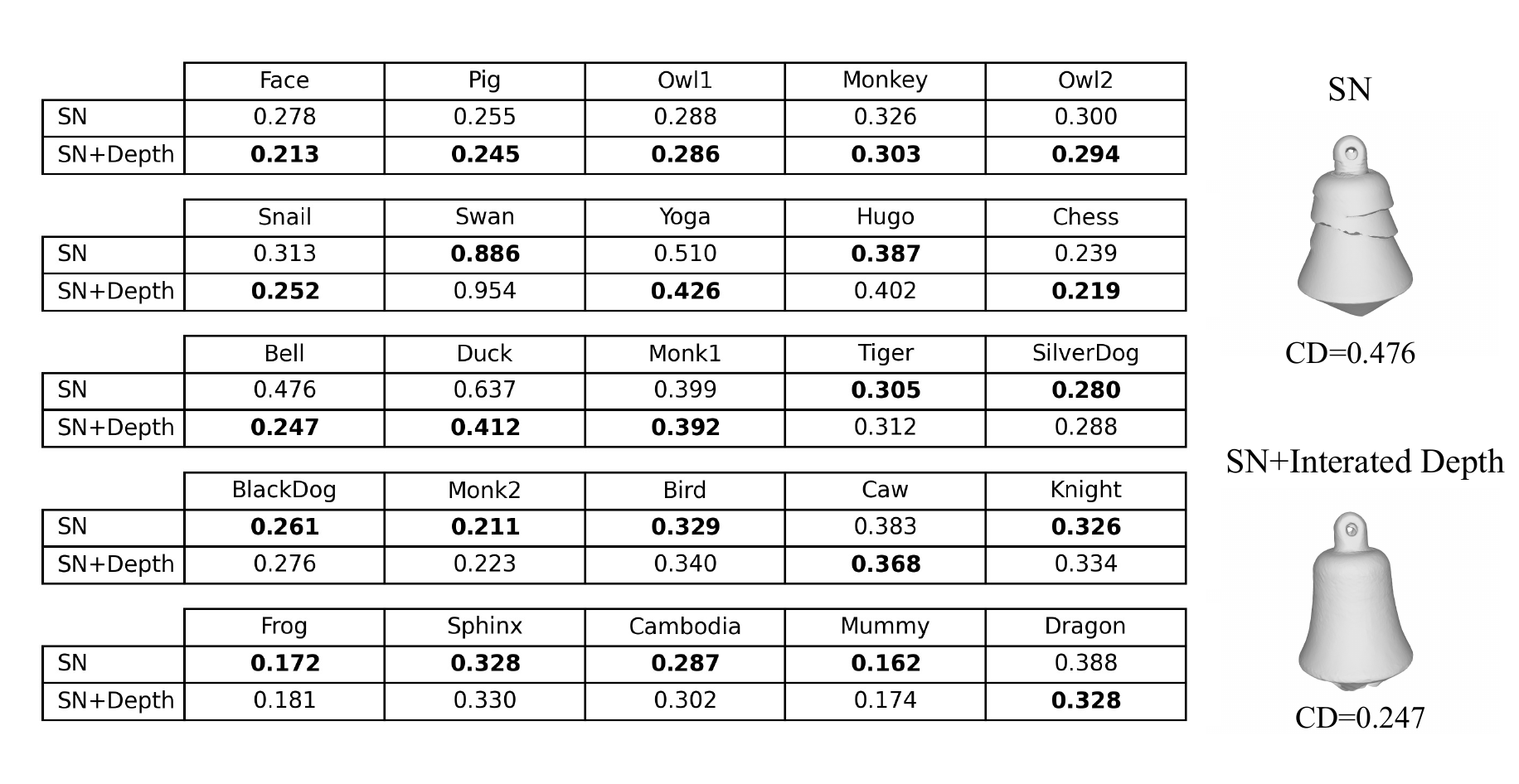}
	\caption{\supernormal with integrated depth improves reconstructions considering per-view shape continuity.}
	\label{fig:acc_eval}
\end{figure}

As shown in \fref{fig:time_eval}, \gaussianS is the fastest method among the baselines but lacks accuracy. \neuss delivers higher reconstruction quality than \gaussianS, albeit with a slower runtime. 
\supernormal is the second fastest and maintains high-quality shape reconstruction, thanks to its patch-based ray marching strategy. Leveraging hash encoding, \vminer and \pisr achieve faster speeds than \wildlight and \nersp, which rely on positional encoding. Overall, \supernormal provides a relatively optimal trade-off between computational efficiency and shape recovery quality. Note that the time for estimating normal maps is not included in this comparison.

% \subsection{Enhancing \supernormal with Integrated depth}

\subsection{Enhancing SuperNormal with Integrated depth}

As shown in \fref{fig:normal_aop_1}, \supernormal produces poor results when the input normal maps are inaccurate. To solve this problem, we incorporate per-view integrated depth as a geometric continuity regularization to improve reconstruction quality. Specifically, we apply a normal integration method~\cite{cao2021normal} to get integrated depth map $\hat{\V{d}}_i$ from surface normal map of the $i$-th view. Given depth map $\V{d}_i$ rendered by SDF, we calculate the inherent scale ambiguity $s_i$ based on least squares, \ie, \hbox{$s_i =  (\V{d}_i^T  \V{d}_i) / (\V{d}_i^T  \hat{\V{d}}_i)$}. Given this scale, we regularize the SDF network in \supernormal with an additional loss:
\begin{eqnarray}
	\mathcal{L}_{depth} = \sum_{i=1}^{N} \left| \hat{\V{d}}_i {s_i} - \V{d}_i \right|.
\end{eqnarray}
As shown in \fref{fig:acc_eval}, incorporating depth regularization improves the mesh reconstruction quality of \supernormal~(average CD decreases from 0.349 to 0.324). The integrated depth maps are continuous, preventing inaccurate discontinuity in recovered mesh.

\section{Conclusion}
We introduce EvalMVX, a real-world dataset designed to simultaneously evaluate MVS, MVPS, and MVSfP methods for surface shape recovery. Our benchmark results reveal the working range and performance of different setups across diverse shapes and reflectances. Notably, MVPS methods achieve the highest shape recovery accuracy due to the additional information provided by illumination variations, which can be conveniently obtained through flash/no-flash data capture. For accurate MVS-based shape reconstruction, \neuss or \nero should be selected based on the reflectivity of the surface. MVSfP methods, however, show lower reconstruction quality than MVS and MVPS, primarily due to the limitations in polarization image quality from snapshot polarization cameras. Based on these findings, we identify several open problems for advancing MVX methods.

\vspace{- 0.2cm}

\paragraph{Open Problems.}

\textbf{1. Limitations of 3DGS Representation.} While 3DGS-based MVS offers high efficiency, its representation remains suboptimal compared to SDF due to geometric inconsistencies between 3D Gaussians and surface normals.
\textbf{2. Reliability of Geometric Priors.} Surface normals estimated from learning-based photometric stereo (MVPS) and polarization cues (MVSfP) provide valuable geometric priors but are susceptible to significant errors stemming from domain gaps or sensor noise. Incorporating multiview consistency and leveraging the degree of polarization as a confidence measure could improve the robustness.
\textbf{3. Efficiency of Normal Acquisition.} Although MVPS methods have shown strong performance, the surface normals produced by off-the-shelf universal photometric stereo approaches~\cite{sdm,hardy2024uni} are computationally expensive in terms of time and memory. Developing lightweight and accurate surface normal acquisition methods is essential for enhancing the scalability and applicability of MVPS.

\vspace{- 0.2cm}

\paragraph{Limitation and future works.} 
Our dataset evaluates 13 MVX methods across 25 objects. Given the rapid development in neural surface reconstruction, it is impractical to include every newly proposed method. To address this, we plan to develop an online evaluation platform to provide continuously updated benchmark results.
Furthermore, as MVX encompasses more than just MVS, MVPS, and MVSfP, we aim to incorporate data from depth and infrared (IR) cameras and extend our dataset scale in future versions to enable broader multiview shape reconstruction evaluations beyond RGB inputs.

\clearpage
\setcounter{page}{1}
\maketitlesupplementary

In this supplementary material, we first provide detailed information on the capture setup and the calibration procedure for the LED light source positions. We also include the viewpoint distribution of each object to offer a better understanding of the multi-view capture settings. We then describe the complete pipeline for ground truth (GT) shape alignment, including optimization runtime and resulting alignment errors. We then show the complexity of selected objects in \ours. Next, we specify the implementation details of the evaluated algorithms and include additional experiments conducted with 10 input views. We further evaluate the reconstructed meshes using mean angular error (MAE) as a complementary metric. We also report additional baseline results for further comparison. Finally, we provide the complete benchmark results containing recovered meshes and error maps reconstructed by each method on \ours.

\section{Capture settings}
\subsection{Obtaining Unpolarized RGB Images}
To capture \ours, we utilized a Lucid Triton RGB polarization camera equipped with a 16mm lens, obtaining raw images in Bayer format with a resolution of $2448 \times 2048$ and an exposure time of 40ms. First, the raw images were demosaiced into four single-channel polarization images: $I_{0}$, $I_{45}$, $I_{90}$, and $I_{135}$, corresponding to polarization angles of \(0^\circ\), \(45^\circ\), \(90^\circ\), and \(135^\circ\), respectively. Each image has a resolution of $1224 \times 1024$. Subsequently, we averaged the four polarization images (after replicating them across RGB channels) to obtain the depolarized RGB image, also at a resolution of $1224 \times 1024$. Finally, a gamma correction with a gamma value of 2.2 was applied to the depolarized image to adjust their brightness and contrast, generating the final RGB images.

\subsection{Viewpoints Distribution}
\begin{figure*}
        \centering 
	\includegraphics[width=\textwidth]{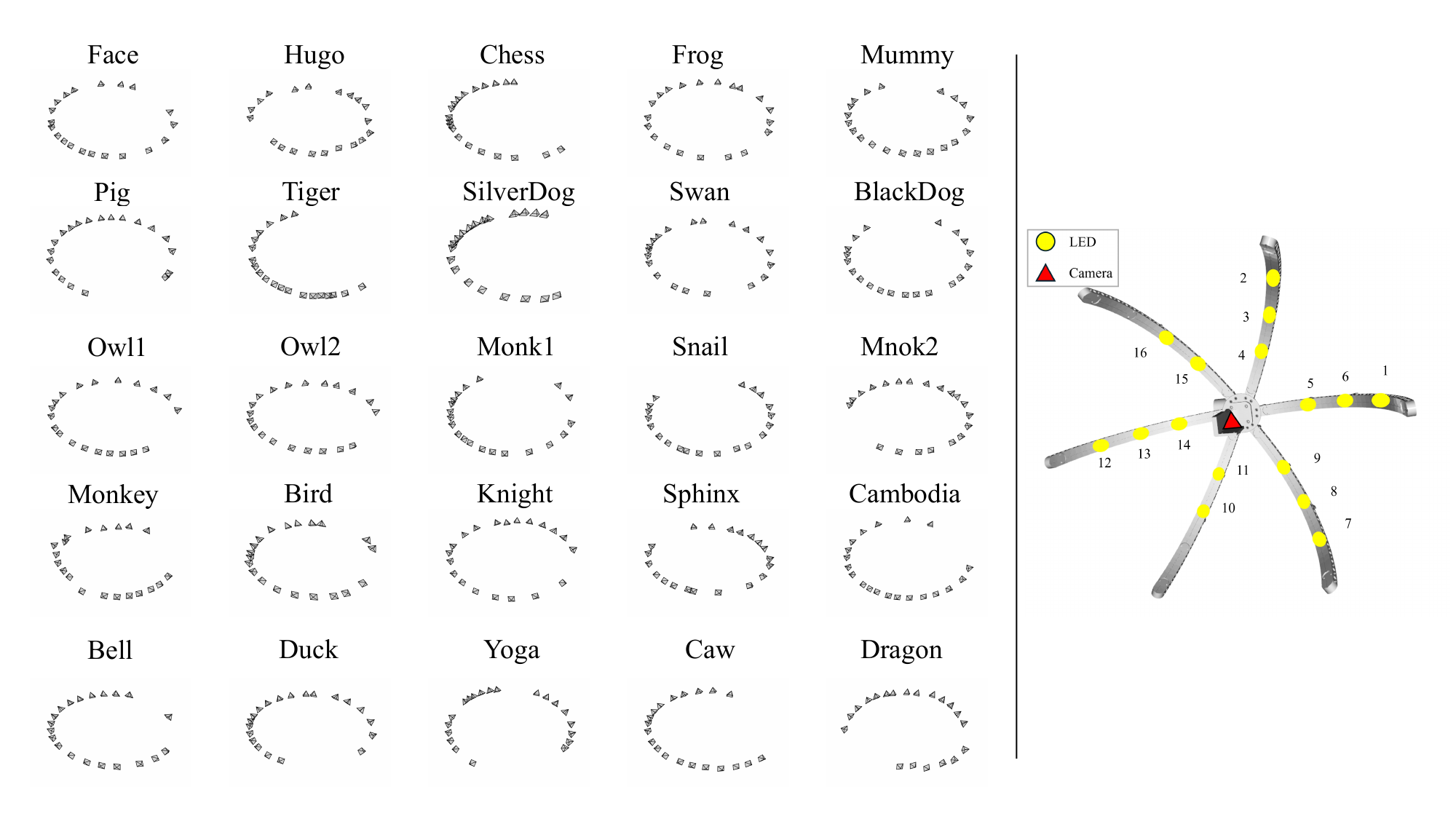}
	\caption{(\textbf{Left}) Distribution of camera poses. (\textbf{Right}) Relative positions between the LEDs and camera.}
	\label{fig:camera_dis}
\end{figure*}

% 添加详细的捕获说明以及相机视角分布和灯光分布，以及和物体中心之间的距离
% 一定要注意合理的表述，不要给自己挖坑
The object is placed at the center of a circular platform and remains static during data capture. The camera is placed above the object at a slanted position and moves horizontally around the platform to capture images from multiple viewpoints. 
The distance between the camera and the object is maintained at a constant value of 0.6--0.8\,m throughout the rotation.
\fref{fig:camera_dis} illustrates the detailed distribution of camera viewpoints across the 25 objects.

\subsection{Light Calibration}
As shown in \fref{fig:camera_dis}, the relative positions between the 16 LED lights and the camera remain fixed during image capture. The camera’s rotation matrix and translation vector in the world coordinate system are obtained through camera calibration. Based on these, the LED positions in the world coordinate system can be computed from the known camera poses and the fixed relative positions between the camera and the LEDs. Specifically, according to the CAD design of our LED rig, the LED coordinates $(x, y, z)$ are defined in the camera coordinate system, where the optical center of the camera is assumed to be the origin. Using the calibrated camera pose, a projection matrix \( \mathbf{P} = [\mathbf{R} \ \mathbf{t}] \), composed of the rotation matrix \( \mathbf{R} \) and the translation vector \( \mathbf{t} \), is used to transform each LED’s coordinates to the world coordinate system. The coordinates of each LED in the world coordinate system \( (X, Y, Z) \) can be obtained via:
\begin{eqnarray}
\begin{bmatrix} X \\ Y \\ Z \end{bmatrix} = \mathbf{R} \cdot \begin{bmatrix} x \\ y \\ z \end{bmatrix} + \mathbf{t}
\end{eqnarray}

\section{GT Shape Alignment}
\begin{figure*}
	\includegraphics[width=\linewidth]{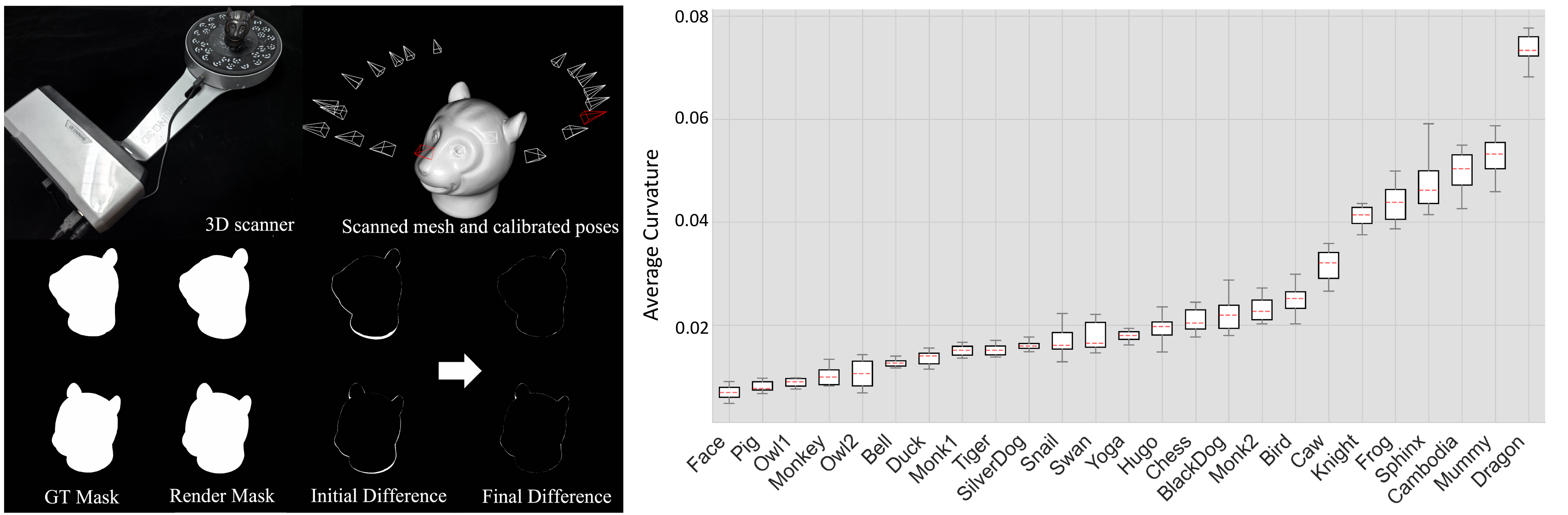}
	\caption{(\textbf{Left}) We align scanned mesh to camera views guided by GT masks. (\textbf{Right}) Curvature distribution of \ours calculated from GT normals quantifies the shape complexity of each object.} 
	\label{fig:curvature_visual}
\end{figure*}

% 补充完善正文中没说清楚的标定设置，正文中的图略显简单
As shown in \fref{fig:curvature_visual}, we scan all objects by EinScan-SP scanner. To minimize scanning errors caused by occlusions or concavities that are difficult to capture from a single viewpoint, each object was scanned in both vertical and horizontal orientations, with multiple rotations to cover all parts.
% 加上对齐程序的loss以及对齐的时间分析，进一步增强说服力
\begin{table*}[h]
\centering
\begin{tabular}{|l|l|l|l|}
\hline
Object & Resolution & Per-view Optimization Time & Average IoU \\ \hline
Pig  & 153×128 $\rightarrow$ 1224×1024 & 6s $\rightarrow$ 2min & 0.9910 $\rightarrow$ 0.9917 \\ \noalign{\hrule}
Cambodia & 153×128 $\rightarrow$ 1224×1024 & 30s $\rightarrow$ 20min & 0.9882 $\rightarrow$ 0.9891 \\ \noalign{\hrule}
\end{tabular}
\caption{Comparison of mesh alignment under different resolutions.}
\label{tab:resolution_down}
\end{table*}

Our camera pose optimization, based on \pytorchddd, minimizes the discrepancy between rendered silhouette masks and the corresponding captured masks. To speed up the process, we reduce the number of faces in the GT mesh and downsample input images to a resolution of $153 \times 128$. As shown in \Tref{tab:resolution_down}, this resolution reduction leads to only marginal drops in alignment accuracy (measured by the average IoU over 20 views) while significantly reducing optimization time. Thus, downsampling proves both effective and efficient. Additionally, we carefully inspected and manually corrected errors in the SAM-generated masks, as their quality greatly impacts the final results.

The optimization time depends on the object’s complexity, with all experiments performed on a single NVIDIA RTX 4090 GPU. For simple objects like {\sc Face} and {\sc Duck}, optimization for each view takes around 10 seconds, while for complex objects like {\sc Cambodia} and {\sc Dragon}, it increases to approximately 30 seconds. We use the binary cross-entropy (BCE) loss function, with final loss values converging between 0.0004 and 0.0006. As shown in \fref{fig:curvature_visual}, the rendered silhouettes align well with the corresponding GT masks, demonstrating the effectiveness of our multi-view silhouette-based alignment approach. After the alignment, we use Blender to inspect the results and render surface normals.

\section{Shape complexity}

As shown in \fref{fig:curvature_visual}, we quantify the shape complexity across different objects in \ours, using average curvature~\cite{heep2024adaptive} calculated on surface normal maps as the metric.

\section{More details of benchmark evaluations}
\subsection{Experiment setup}
% 申明使用的算法实现版本以及执行的轮次，补充说明执行时间的评估依据
We use a single NVIDIA GTX 4090 GPU for each method to run the benchmark evaluations. 
% 我们把算法实现的声明移动到此处，确实没什么重要贡献
Notably, for \rnbneus, we adopt the faster implementation RnbNeuS2, which is the official CUDA implementation of \rnbneus. For fair comparison with \supernormal, we use only normal maps as input. For \gaussianS, we experimentally found that using only RGB images as input yields better reconstruction results compared to using both the predicted normal maps from \unips and the RGB images. Therefore, we report the reconstruction results obtained using RGB images only. Moreover, we observed that \petneus failed to produce valid reconstructions under the standard input setting. To ensure the method could yield usable results and allow for meaningful comparison, we manually increased the illumination intensity in the ambient input images. This adjustment was applied only to \petneus and not to other methods.
Except for that, all other methods are evaluated using their official implementations.

% 补充说明算法执行的时间
Since different algorithms converge at different rates, we set varying numbers of training iterations for each method accordingly. Specifically, \neuss, \supernormal, and \rnbneus are trained for 30,000 iterations; \nero, \petneus and \wildlight are trained for 100,000 iterations; \gaussianS is trained for 15,000 iterations; \vminer is trained for a total of 40,000 iterations; \mvas is trained for 50 epochs; \nersp is trained for 50,000 iterations; and \pisr is trained for 20,000 iterations.

\subsection{Sparse view evaluation}
% 尽量不要让读者混淆，声明我们除了刻意提到以外，所有的实验结果都是20个视角下的
To evaluate the impact of reduced viewpoint numbers on the reconstruction performance of different MVX methods, we selected 10 uniformly spaced viewpoints from the original set of 20. We then evaluated the reconstruction quality of each method under this 10-view setting. The results are shown in \Tref{tab:benchmark_cd_10}. We observed that \supernormal and \rnbneus, which rely on normal maps for surface reconstruction, remained robust even when using only half of the input views. This is likely because normal maps offer dense geometric details and enforce consistency across the surface, helping maintain smooth and accurate shapes with limited observations.

\begin{table*}
    \caption{Benchmark results on \ours with 10 views - CD evaluation in millimeter over all 25 objects, where the smallest and second smallest ones are shown in bold and underlined, respectively.}
    \centering
    \resizebox{\textwidth}{!}{
    \begin{tabular}{c|cccc|cccc|ccc}
        \toprule
        \multirow{2}{*}{Method} &  \multicolumn{4}{c|}{\cellcolor{MyPink} MVS} & \multicolumn{4}{c|}{\cellcolor{MyPeach} MVPS} & \multicolumn{3}{c}{\cellcolor{MyLightBlue} MVSfP} \\
        \cmidrule{2-12}
        & \neuss & \nero & \petneus & \gaussianS & \rnbneus & \supernormal & \vminer & \wildlight & \mvas & \nersp & \pisr \\
        \midrule
        {\sc Face} & \textbf{0.254} & 0.281 & 0.303 & 0.670 & 0.541 & \underline{0.260} & 0.370 & 0.366 & 0.420 & 0.588 & 0.396 \\
        {\sc Pig} & 0.324 & \underline{0.279} & 0.461 & 0.705 & 0.386 & \textbf{0.254} & 0.456 & 0.360 & 0.498 & 0.424 & 0.355 \\
        {\sc Owl1} & 0.352 & \underline{0.335} & 0.423 & 0.759 & 0.424 & \textbf{0.312} & 0.541 & 0.532 & 0.419 & 0.424 & 0.439 \\
        {\sc Monkey} & 0.561 & 0.346 & 0.698 & 0.736 & 0.425 & \textbf{0.308} & 0.457 & 0.441 & 0.428 & 0.514 & \underline{0.337} \\
        {\sc Owl2} & \underline{0.391} & 0.410 & 0.478 & 0.725 & 0.503 & \textbf{0.315} & 0.614 & 0.513 & 0.552 & 0.490 & 0.444 \\
        {\sc Bell} & 0.319 & \textbf{0.232} & 1.372 & 0.736 & 0.605 & 0.512 & 0.489 & 0.303 & 0.476 & 0.438 & \underline{0.302} \\
        {\sc Duck} & 0.471 & \textbf{0.258} & 0.747 & 0.767 & 0.706 & 0.629 & 0.566 & 0.450 & 0.604 & 0.676 & \underline{0.301} \\
        {\sc Monk1} & 0.658 & \underline{0.423} & 0.661 & 1.341 & 0.501 & \textbf{0.395} & 0.866 & 0.607 & 0.999 & 0.891 & 0.543 \\
        {\sc Tiger} & 0.400 & 0.383 & 0.494 & 0.772 & \underline{0.353} & \textbf{0.314} & 0.529 & 0.492 & 0.642 & 0.503 & 0.467 \\
        {\sc SilverDog} & 0.351 & 0.368 & 0.546 & 0.683 & 0.357 & \textbf{0.290} & 0.438 & 0.374 & 0.749 & 0.426 & \underline{0.318} \\
        {\sc Snail} & 0.334 & \textbf{0.274} & 0.424 & 0.538 & 0.349 & \underline{0.311} & 0.415 & 0.403 & 0.569 & 0.375 & 0.360 \\
        {\sc Swan} & 1.147 & 1.156 & 1.254 & 1.963 & \textbf{0.954} & \underline{1.053} & 1.837 & 1.339 & 1.457 & 1.249 & 1.353 \\
        {\sc Yoga} & 0.474 & \textbf{0.303} & 0.999 & 0.741 & 0.522 & 0.563 & 0.706 & \underline{0.453} & 0.570 & 0.612 & 0.457 \\
        {\sc Hugo} & 0.553 & 0.512 & 0.563 & 0.769 & \underline{0.437} & \textbf{0.407} & 0.668 & 0.728 & 0.773 & 0.677 & 0.653 \\
        {\sc Chess} & \textbf{0.255} & 0.266 & 0.308 & 0.648 & 0.293 & \underline{0.257} & 0.384 & 0.334 & 0.550 & 0.478 & 0.361 \\
        {\sc BlackDog} & 0.356 & 0.359 & N/A & 0.700 & \underline{0.335} & \textbf{0.276} & 0.404 & 0.423 & 0.581 & 0.392 & 0.382 \\
        {\sc Monk2} & 0.440 & 0.443 & 0.463 & 0.671 & \underline{0.386} & \textbf{0.227} & 0.516 & 0.467 & 0.607 & 0.497 & 0.416 \\
        {\sc Bird} & 0.486 & \textbf{0.299} & 1.163 & 0.585 & \underline{0.332} & 0.386 & 0.705 & 0.595 & 0.604 & 0.452 & 0.510 \\
        {\sc Caw} & 0.541 & \textbf{0.355} & 0.500 & 0.634 & 0.411 & \underline{0.359} & 0.885 & 0.673 & 1.000 & 0.673 & 0.405 \\
        {\sc Knight} & 0.404 & 0.404 & 1.497 & 0.558 & \underline{0.344} & \textbf{0.339} & 0.467 & 0.485 & 0.650 & 0.575 & 0.451 \\
        {\sc Frog} & 0.321 & 0.346 & 0.326 & 0.451 & \underline{0.228} & \textbf{0.175} & 0.412 & 0.435 & 0.532 & 0.392 & 0.428 \\
        {\sc Sphinx} & 0.624 & 0.618 & 0.782 & 0.862 & \underline{0.368} & \textbf{0.351} & 0.776 & 0.897 & 0.913 & 0.788 & 0.731 \\
        {\sc Cambodia} & 0.545 & 0.616 & 0.602 & 0.640 & \underline{0.315} & \textbf{0.311} & 0.894 & 0.727 & 0.876 & 0.742 & 0.588 \\
        {\sc Mummy} & 0.291 & 0.305 & 0.482 & 0.456 & \underline{0.176} & \textbf{0.173} & 0.333 & 0.295 & 0.466 & 0.324 & 0.309 \\
        {\sc Dragon} & 0.492 & 0.454 & N/A & 0.492 & \textbf{0.400} & 0.457 & 0.645 & 0.515 & 0.701 & 0.578 & \underline{0.441} \\
        Average & \cellcolor{gray!20} 0.454 & \cellcolor{gray!20} \underline{0.401} & \cellcolor{gray!20} 0.676 & \cellcolor{gray!20} 0.744 & \cellcolor{gray!20} 0.426 & \cellcolor{gray!20} \textbf{0.369} & \cellcolor{gray!20} 0.615 & \cellcolor{gray!20} 0.528 & \cellcolor{gray!20} 0.665 & \cellcolor{gray!20} 0.567 & \cellcolor{gray!20} 0.470 \\
        Median & \cellcolor{gray!20} 0.404 & \cellcolor{gray!20} \underline{0.355} & \cellcolor{gray!20} 0.546 & \cellcolor{gray!20} 0.700 & \cellcolor{gray!20} 0.386 & \cellcolor{gray!20} \textbf{0.314} & \cellcolor{gray!20} 0.529 & \cellcolor{gray!20} 0.467 & \cellcolor{gray!20} 0.604 & \cellcolor{gray!20} 0.503 & \cellcolor{gray!20} 0.428 \\
        \bottomrule
    \end{tabular}
    }  % 结束 resizebox
    \label{tab:benchmark_cd_10}
\end{table*}

\subsection{Surface normal evaluation}
% 此处仅展示20个视角的法线MAE即可
To further assess surface detail recovery, we compute the MAE between the ground truth and recovered surface normals across all 20 views, and report the average MAE for each object. As shown in \Tref{tab:benchmark_mae}, these results are consistent with those evaluated based on Chamfer Distance (CD). However, we observed that the evaluation may not be entirely fair to \supernormal and \rnbneus, as they benefit from the accurate normals provided by \unips, and continuously optimize the surface normals under multi-view constraints. Consequently, they achieve the lowest or second-lowest MAE for most objects. 

\begin{table*}
    \caption{Benchmark results on \ours - MAE evaluation in degree over all 25 objects, where the smallest and second smallest ones are shown in bold and underlined, respectively.}
    \centering
    \resizebox{\textwidth}{!}{
    \begin{tabular}{c|cccc|cccc|ccc}
        \toprule
        \multirow{2}{*}{Method} &  \multicolumn{4}{c|}{\cellcolor{MyPink} MVS} & \multicolumn{4}{c|}{\cellcolor{MyPeach} MVPS} & \multicolumn{3}{c}{\cellcolor{MyLightBlue} MVSfP} \\
        \cmidrule{2-12}
        & \neuss & \nero & \petneus & \gaussianS & \rnbneus & \supernormal & \vminer & \wildlight & \mvas & \nersp & \pisr \\
        \midrule
        {\sc Face} & 5.96 & \underline{3.36} & 4.88 & 12.32 & 4.19 & \textbf{2.22} & 6.92 & 3.95 & 8.09 & 7.52 & 14.98 \\
        {\sc Pig} & 9.57 & \underline{4.58} & 8.84 & 16.53 & 5.62 & \textbf{4.15} & 9.40 & 5.06 & 8.86 & 9.60 & 8.69 \\
        {\sc Owl1} & 7.71 & \underline{5.17} & 12.84 & 12.90 & 5.46 & \textbf{3.35} & 7.28 & 7.00 & 6.58 & 7.51 & 9.51 \\
        {\sc Monkey} & 16.64 & \underline{5.62} & 25.82 & 18.51 & 6.20 & \textbf{4.41} & 17.89 & 7.08 & 8.42 & 13.05 & 7.60 \\
        {\sc Owl2} & 12.35 & \underline{6.36} & 11.90 & 17.73 & 7.24 & \textbf{3.98} & 10.64 & 9.16 & 9.97 & 10.28 & 11.15 \\
        {\sc Bell} & 12.91 & \underline{5.46} & 23.41 & 16.95 & 10.55 & 10.32 & 15.12 & \textbf{5.36} & 12.83 & 9.41 & 14.56 \\
        {\sc Duck} & 16.94 & \textbf{6.67} & 20.41 & 22.45 & 10.96 & 11.35 & 15.52 & 8.20 & 12.93 & 24.43 & 11.94 \\
        {\sc Monk1} & 13.34 & 9.48 & 14.75 & 22.97 & \underline{6.47} & \textbf{5.59} & 16.55 & 12.96 & 17.06 & 18.60 & 13.35 \\
        {\sc Tiger} & 11.75 & 8.12 & 13.46 & 22.69 & \underline{6.03} & \textbf{4.83} & 14.64 & 8.89 & 14.05 & 11.59 & 13.74 \\
        {\sc SilverDog} & 13.66 & 7.40 & 14.81 & 21.95 & 7.00 & 5.00 & 22.96 & 9.39 & 14.60 & 10.51 & 10.26 \\
        {\sc Snail} & 13.99 & \underline{8.24} & 13.50 & 22.06 & 8.27 & \textbf{6.85} & 16.19 & 9.35 & 16.53 & 11.62 & 15.58 \\
        {\sc Swan} & 23.18 & 23.75 & 11.69 & 27.13 & \underline{9.19} & \textbf{8.11} & 23.92 & 23.25 & 25.74 & 16.53 & 24.27 \\
        {\sc Yoga} & 20.97 & \textbf{6.64} & 29.10 & 27.73 & 19.89 & 20.78 & 17.03 & \underline{8.73} & 22.07 & 19.29 & 17.77 \\
        {\sc Hugo} & 15.13 & 12.84 & 13.68 & 19.36 & \underline{7.32} & \textbf{5.93} & 15.04 & 15.46 & 18.76 & 16.16 & 22.27 \\
        {\sc Chess} & 10.23 & 8.14 & 8.83 & 18.25 & \underline{7.24} & \textbf{5.92} & 13.34 & 9.64 & 17.74 & 12.70 & 16.83 \\
        {\sc BlackDog} & 15.50 & 13.82 & 15.96 & 22.12 & \underline{10.31} & \textbf{8.12} & 16.18 & 12.30 & 18.16 & 13.44 & 15.46 \\
        {\sc Monk2} & 14.78 & 12.78 & 11.81 & 22.47 & \underline{9.48} & \textbf{6.67} & 15.72 & 14.63 & 17.49 & 15.00 & 16.33 \\
        {\sc Bird} & 24.50 & 13.20 & 36.37 & 27.96 & 19.69 & 25.68 & 21.97 & \textbf{10.52} & 21.52 & 18.38 & 20.41 \\
        {\sc Caw} & 16.94 & \textbf{14.12} & 20.73 & 26.63 & \underline{14.79} & 17.48 & 22.91 & 15.22 & 32.12 & 25.70 & 18.04 \\
        {\sc Knight} & 20.78 & 17.83 & 25.41 & 27.51 & \textbf{12.84} & \underline{13.24} & 22.26 & 20.80 & 25.29 & 22.10 & 21.73 \\
        {\sc Frog} & 19.31 & 20.32 & 16.67 & 25.50 & \underline{10.55} & \textbf{8.78} & 22.57 & 20.38 & 26.34 & 22.05 & 26.89 \\
        {\sc Sphinx} & 24.87 & 25.35 & 28.67 & 30.00 & \textbf{17.59} & \underline{17.68} & 33.65 & 25.40 & 31.24 & 29.83 & 30.06 \\
        {\sc Cambodia} & 26.80 & 27.56 & 25.45 & 30.15 & \textbf{15.24} & \underline{17.84} & 30.87 & 27.30 & 32.44 & 29.64 & 26.87 \\
        {\sc Mummy} & 22.19 & 21.59 & 18.54 & 27.32 & \underline{14.58} & \textbf{14.16} & 22.93 & 20.95 & 27.82 & 21.60 & 23.96 \\
        {\sc Dragon} & 33.78 & 32.70 & 35.53 & 37.25 & \textbf{29.64} & 32.10 & 35.40 & 33.06 & 40.72 & 34.97 & 34.21 \\
        Average & \cellcolor{gray!20} 16.95 & \cellcolor{gray!20} 12.84 & \cellcolor{gray!20} 18.52 & \cellcolor{gray!20} 23.06 & \cellcolor{gray!20} \underline{11.05} & \cellcolor{gray!20} \textbf{10.58} & \cellcolor{gray!20} 18.68 & \cellcolor{gray!20} 13.76 & \cellcolor{gray!20} 19.49 & \cellcolor{gray!20} 17.26 & \cellcolor{gray!20} 17.86 \\
        Median & \cellcolor{gray!20} 15.50 & \cellcolor{gray!20} \underline{9.48} & \cellcolor{gray!20} 15.96 & \cellcolor{gray!20} 22.47 & \cellcolor{gray!20} \underline{9.48} & \cellcolor{gray!20} \textbf{8.11} & \cellcolor{gray!20} 16.55 & \cellcolor{gray!20} 10.52 & \cellcolor{gray!20} 17.74 & \cellcolor{gray!20} 16.16 & \cellcolor{gray!20} 16.33 \\
        \bottomrule
    \end{tabular}
    }  % 结束 resizebox
    \label{tab:benchmark_mae}
\end{table*}

\subsection{Additional Baseline Results}
We further evaluate the performance of the classical MVS method \neus and \neudf on \ours using all 20 views, as shown in \Tref{tab:benchmark_cd_neus}. The reconstructed meshes exhibit competitive quality across diverse objects and materials, highlighting \neus as a solid baseline for neural surface reconstruction methods.

\begin{table*}
    \caption{Extended benchmark results on EvalMVX, including the additional \neus and \neudf results.}
    \centering
    \resizebox{\textwidth}{!}{
    \begin{tabular}{c|cccccc|cccc|ccc}
        \toprule
        \multirow{2}{*}{Method} &  \multicolumn{6}{c|}{\cellcolor{MyPink} MVS} & \multicolumn{4}{c|}{\cellcolor{MyPeach} MVPS} & \multicolumn{3}{c}{\cellcolor{MyLightBlue} MVSfP} \\
        \cmidrule{2-14}
        & \neus & \neudf & \neuss & \nero & \petneus & \gaussianS & \rnbneus & \supernormal & \vminer & \wildlight & \mvas & \nersp & \pisr \\
        \midrule
        {\sc Face} & 0.314 & 0.267 & \textbf{0.234} & 0.411 & 0.303 & 0.540 & 0.614 & 0.277 & 0.400 & \underline{0.250} & 0.389 & 0.330 & 0.308 \\
        {\sc Pig} & 0.292 & 0.343 & 0.312 & 0.282 & 0.408 & 0.720 & 0.379 & \textbf{0.256} & 0.470 & \underline{0.265} & 0.446 & 0.447 & 0.286 \\
        {\sc Owl1} & \underline{0.306} & 0.413 & 0.327 & \underline{0.306} & 0.360 & 0.636 & 0.443 & \textbf{0.288} & 0.319 & 0.342 & 0.354 & 0.371 & 0.423 \\
        {\sc Monkey} & 0.548 & 0.862 & 0.487 & 0.345 & 0.664 & 0.646 & 0.393 & \underline{0.325} & 0.530 & 0.335 & 0.365 & 0.501 & \textbf{0.311} \\
        {\sc Owl2} & \underline{0.319} & 0.392 & 0.383 & 0.321 & 0.393 & 0.639 & 0.504 & \textbf{0.300} & 0.415 & 0.377 & 0.444 & 0.405 & 0.390 \\
        {\sc Bell} & 0.438 & 0.729 & 0.321 & \underline{0.254} & 1.255 & 0.635 & 0.575 & 0.476 & 0.399 & \textbf{0.200} & 0.411 & 0.373 & 0.292 \\
        {\sc Duck} & 0.707 & 0.336 & 0.413 & \underline{0.249} & 0.717 & 0.684 & 0.739 & 0.637 & 0.548 & 0.268 & 0.490 & 0.685 & \textbf{0.245} \\
        {\sc Monk1} & \textbf{0.348} & 0.519 & 0.527 & 0.401 & 0.523 & 1.074 & 0.494 & \underline{0.399} & 0.626 & 0.502 & 0.772 & 0.802 & 0.406 \\
        {\sc Tiger} & 0.354 & 0.512 & 0.385 & 0.382 & 0.520 & 0.703 & \underline{0.347} & \textbf{0.305} & 0.498 & 0.359 & 0.559 & 0.454 & 0.429 \\
        {\sc SilverDog} & 0.389 & 0.382 & 0.336 & \underline{0.310} & 0.432 & 0.618 & 0.355 & \textbf{0.280} & 0.555 & 0.324 & 0.520 & 0.363 & 0.313 \\
        {\sc Snail} & 0.329 & 0.345 & 0.309 & \textbf{0.264} & 0.346 & 0.515 & 0.355 & 0.314 & 0.409 & \underline{0.267} & 0.486 & 0.320 & 0.344 \\
        {\sc Swan} & 0.804 & \textbf{0.743} & 1.113 & 1.156 & 1.071 & 2.176 & 0.971 &\underline{0.886} & 1.195 & 1.212 & 1.406 & 1.089 & 1.177 \\
        {\sc Yoga} & 0.599 & 1.114 & 0.452 & \textbf{0.284} & 0.738 & 0.675 & 0.549 & 0.510 & 0.426 & \underline{0.295} & 0.489 & 0.517 & 0.410 \\
        {\sc Hugo} & 0.504 & 0.657 & 0.513 & 0.490 & 0.522 & 0.657 & \underline{0.451} & \textbf{0.387} & 0.543 & 0.543 & 0.646 & 0.572 & 0.598 \\
        {\sc Chess} & 0.309 & 0.286 & \underline{0.248} & 0.262 & 0.266 & 0.507 & 0.286 & \textbf{0.240} & 0.373 & 0.282 & 0.497 & 0.353 & 0.324 \\
        {\sc BlackDog} & \underline{0.300} & 0.601 & 0.337 & 0.393 & 0.428 & 0.597 & 0.322 & \textbf{0.262} & 0.387 & 0.320 & 0.520 & 0.348 & 0.364 \\
        {\sc Monk2} & \underline{0.262} & 0.266 & 0.391 & 0.386 & 0.332 & 0.590 & 0.324 & \textbf{0.211} & 0.448 & 0.377 & 0.473 & 0.373 & 0.383 \\
        {\sc Bird} & 0.831 & 1.262 & 0.447 & 0.396 & 0.727 & 0.552 & \underline{0.319} & 0.330 & 0.781 & \textbf{0.263} & 0.591 & 0.481 & 0.474 \\
        {\sc Caw} & 0.384 & 0.386 & 0.421 & \underline{0.367} & \textbf{0.338} & 0.618 & 0.375 & 0.383 & 0.818 & 0.369 & 0.799 & 0.652 & 0.387 \\
        {\sc Knight} & 0.464 & 0.648 & 0.394 & 0.405 & 0.795 & 0.503 & \underline{0.363} & \textbf{0.326} & 0.534 & 0.408 & 0.566 & 0.469 & 0.462 \\
        {\sc Frog} & 0.261 & 0.288 & 0.270 & 0.339 & 0.289 & 0.385 & \underline{0.238} & \textbf{0.171} & 0.395 & 0.308 & 0.483 & 0.351 & 0.352 \\
        {\sc Sphinx} & 0.497 & 0.755 & 0.525 & 0.546 & 0.517 & 0.699 & \underline{0.355} & \textbf{0.328} & 0.901 & 0.605 & 0.756 & 0.717 & 0.605 \\
        {\sc Cambodia} & 0.377 & 0.554 & 0.460 & 0.522 & 0.406 & 0.545 & \underline{0.291} & \textbf{0.287} & 0.726 & 0.578 & 0.822 & 0.637 & 0.512 \\
        {\sc Mummy} & 0.438 & 0.299 & 0.286 & 0.304 & 0.332 & 0.416 & \underline{0.172} & \textbf{0.162} & 0.269 & 0.265 & 0.436 & 0.296 & 0.283 \\
        {\sc Dragon} & 0.637 & 0.567 & 0.458 & 0.468 & 0.666 & 0.446 & \textbf{0.341} & \underline{0.387} & 0.593 & 0.479 & 0.578 & 0.511 & 0.449 \\
        Average & \cellcolor{gray!20} 0.440 & \cellcolor{gray!20} 0.541 & \cellcolor{gray!20} 0.414 & \cellcolor{gray!20} 0.394 & \cellcolor{gray!20} 0.534 & \cellcolor{gray!20} 0.671 & \cellcolor{gray!20} 0.422 & \cellcolor{gray!20} \textbf{0.349} & \cellcolor{gray!20} 0.542 & \cellcolor{gray!20} \underline{0.392} & \cellcolor{gray!20} 0.572 & \cellcolor{gray!20} 0.497 & \cellcolor{gray!20} 0.421 \\
        Median & \cellcolor{gray!20} 0.384 & \cellcolor{gray!20} 0.512 & \cellcolor{gray!20} 0.391 & \cellcolor{gray!20} 0.367 & \cellcolor{gray!20} 0.432 & \cellcolor{gray!20} 0.618 & \cellcolor{gray!20} 0.363 & \cellcolor{gray!20} \textbf{0.314} & \cellcolor{gray!20} 0.498 & \cellcolor{gray!20} \underline{0.335} & \cellcolor{gray!20} 0.497 & \cellcolor{gray!20} 0.454 & \cellcolor{gray!20} 0.387 \\
        \bottomrule
    \end{tabular}
    }  % 结束 resizebox
    \label{tab:benchmark_cd_neus}
\end{table*}

\subsection{Complete benchmark results}
\begin{figure*}
    \centering 
    \includegraphics[width=\linewidth]{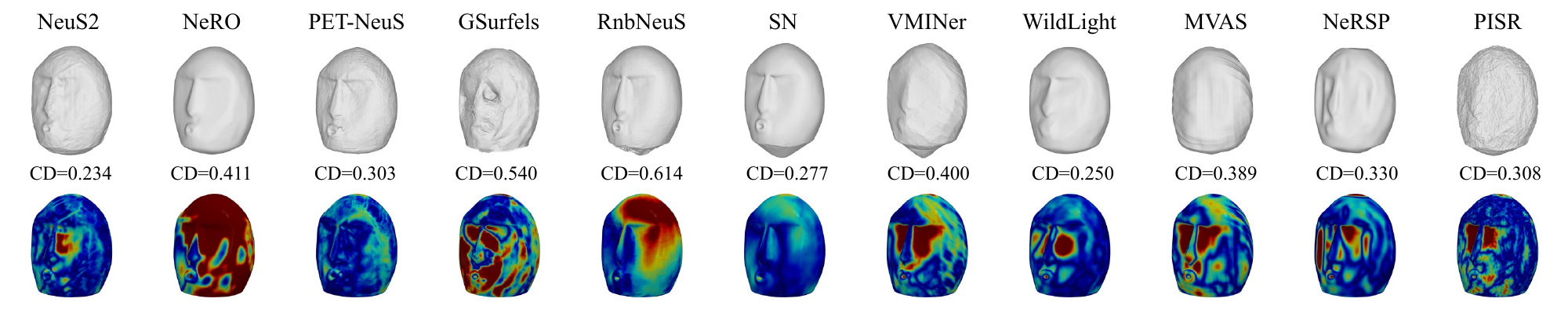}  
    \caption{Recovered shapes and error maps of {\sc Face}.}
    \label{fig:face_result}
\end{figure*}

\begin{figure*}
    \centering 
    \includegraphics[width=\linewidth]{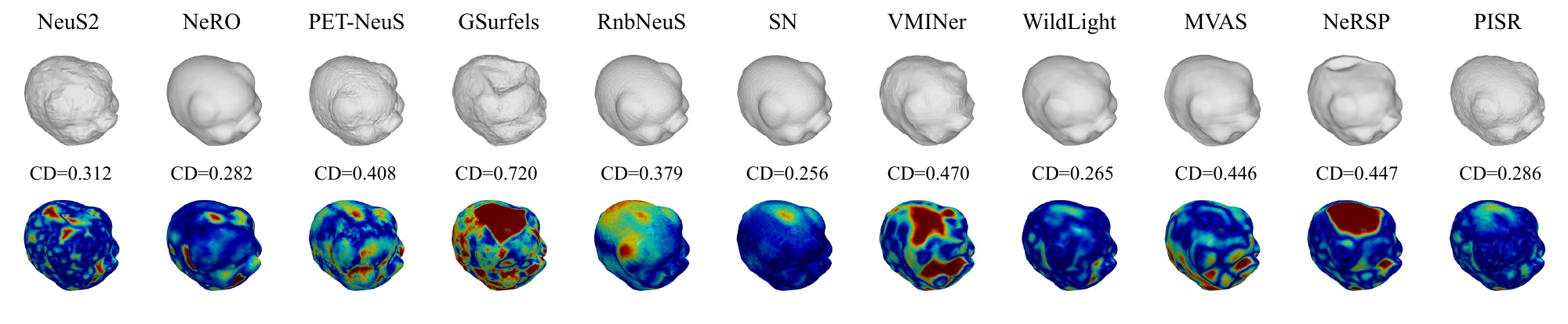}  
    \caption{Recovered shapes and error maps of {\sc Pig}.}
    \label{fig:pig_result}
\end{figure*}

\begin{figure*}
    \centering 
    \includegraphics[width=\linewidth]{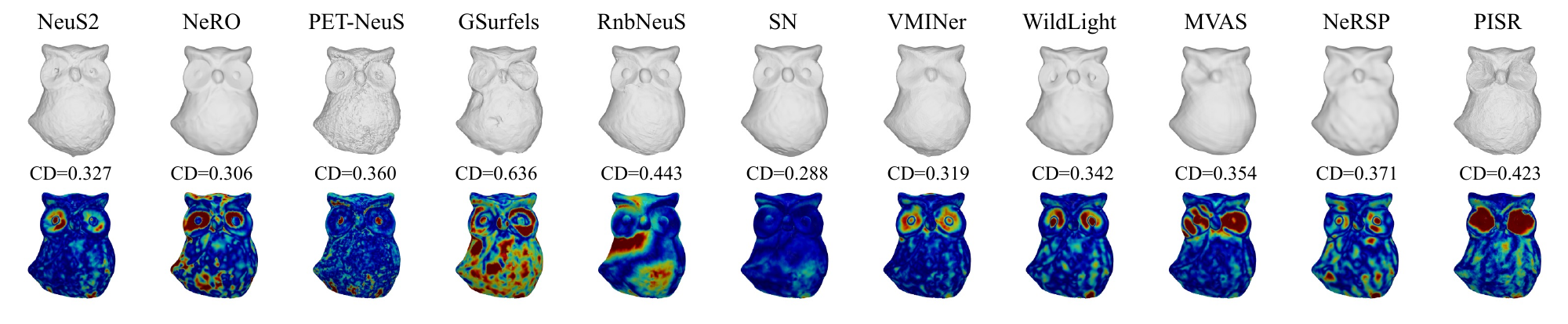}  
    \caption{Recovered shapes and error maps of {\sc Owl1}.}
    \label{fig:owl1_result}
\end{figure*}

\begin{figure*}
    \centering 
    \includegraphics[width=\linewidth]{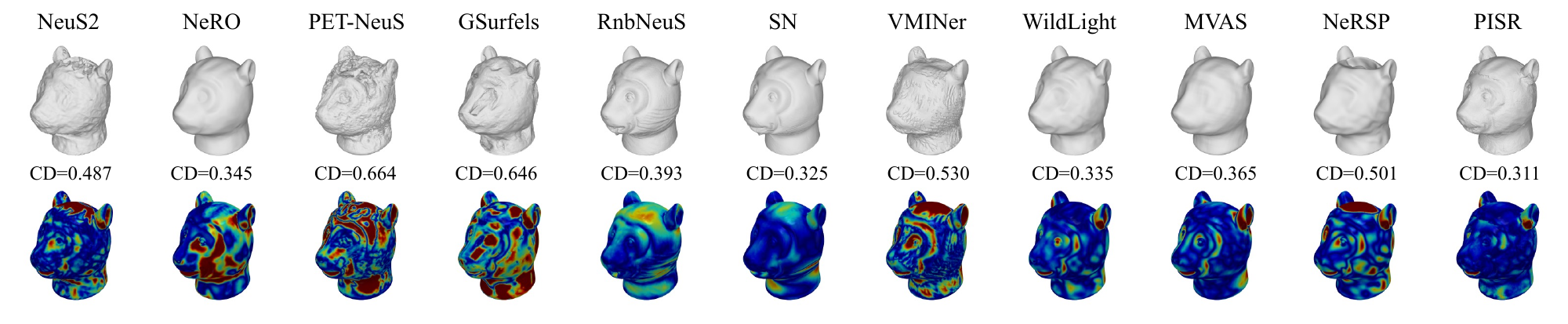}  
    \caption{Recovered shapes and error maps of {\sc Monkey}.}
    \label{fig:monkeyhead_result}
\end{figure*}

\begin{figure*}
    \centering 
    \includegraphics[width=\linewidth]{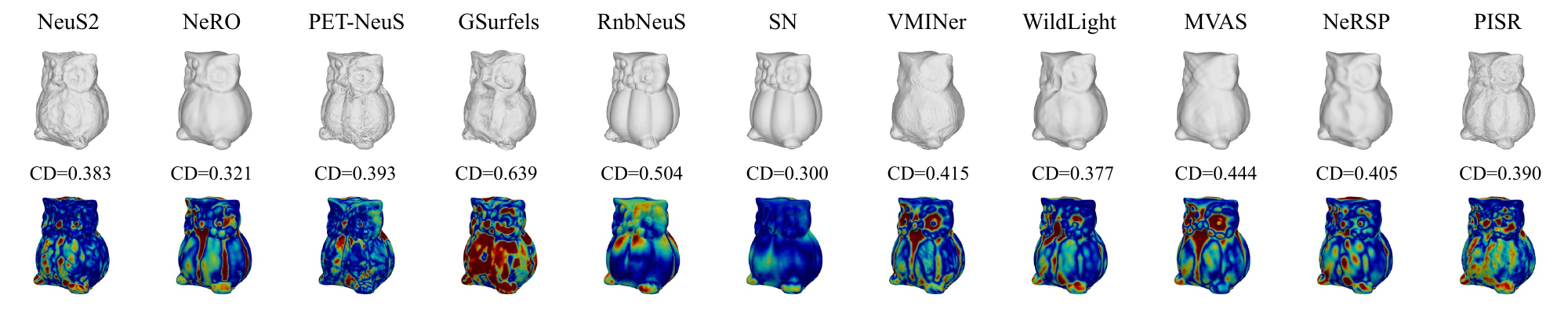}  
    \caption{Recovered shapes and error maps of {\sc Owl2}.}
    \label{fig:owl2_result}
\end{figure*}

\begin{figure*}
    \centering 
    \includegraphics[width=\linewidth]{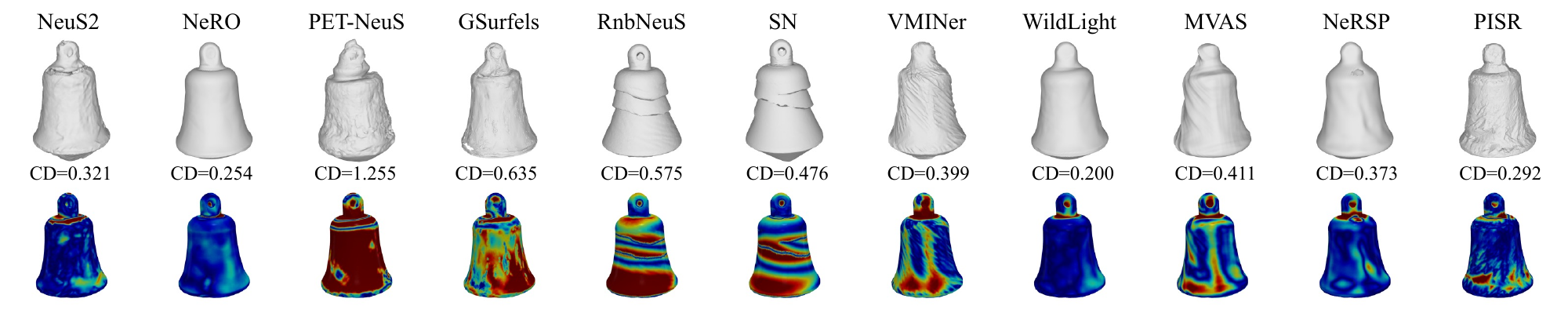}  
    \caption{Recovered shapes and error maps of {\sc Bell}.}
    \label{fig:bell_result}
\end{figure*}

\begin{figure*}
    \centering 
    \includegraphics[width=\linewidth]{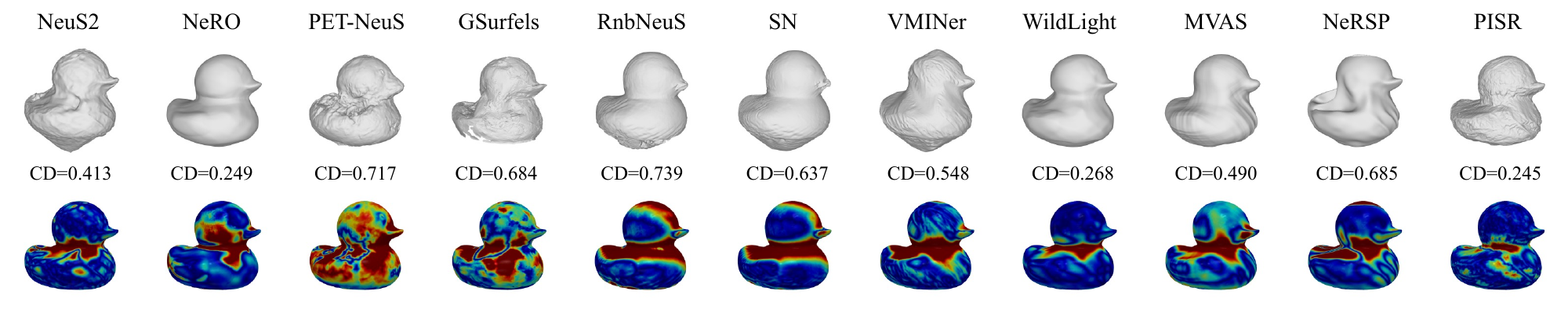}  
    \caption{Recovered shapes and error maps of {\sc Duck}.}
    \label{fig:duck_result}
\end{figure*}

\begin{figure*}
    \centering 
    \includegraphics[width=\linewidth]{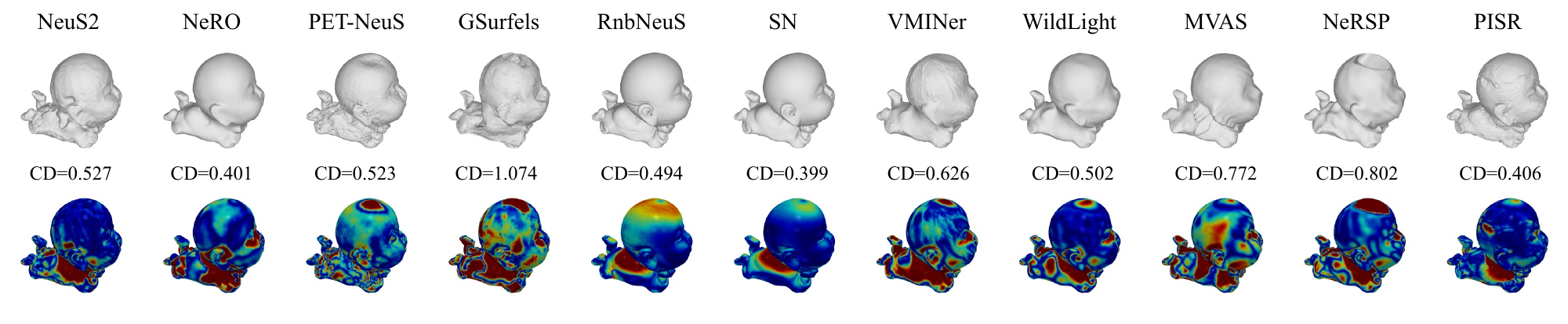}  
    \caption{Recovered shapes and error maps of {\sc Monk1}.}
    \label{fig:monk1_result}
\end{figure*}

\begin{figure*}
    \centering 
    \includegraphics[width=\linewidth]{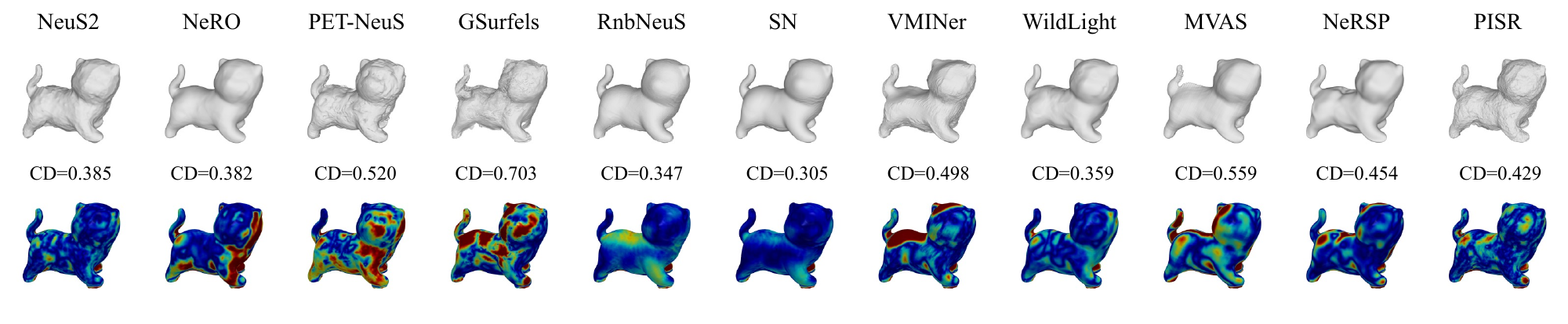}  
    \caption{Recovered shapes and error maps of {\sc Tiger}.}
    \label{fig:tiger_result}
\end{figure*}

\begin{figure*}
    \centering 
    \includegraphics[width=\linewidth]{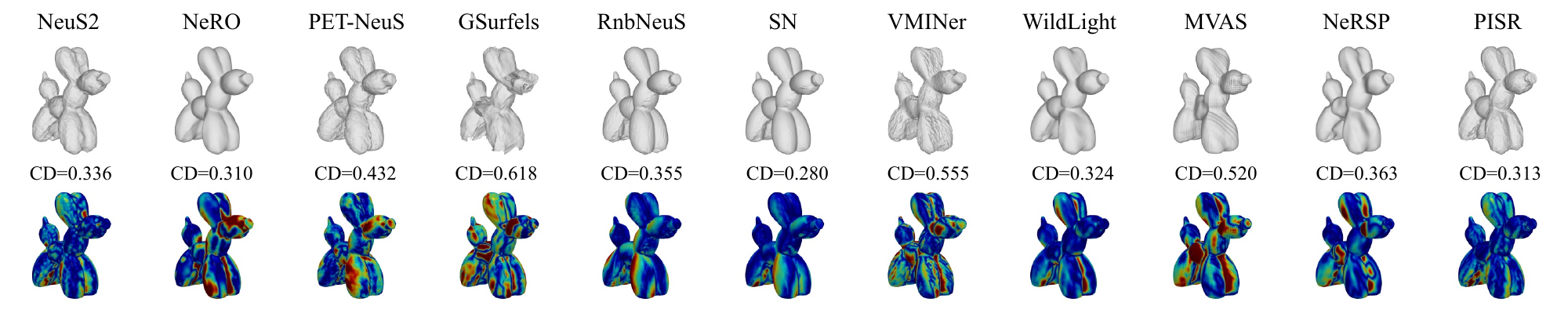}  
    \caption{Recovered shapes and error maps of {\sc SilverDog}.}
    \label{fig:silverdog_result}
\end{figure*}

\begin{figure*}
    \centering 
    \includegraphics[width=\linewidth]{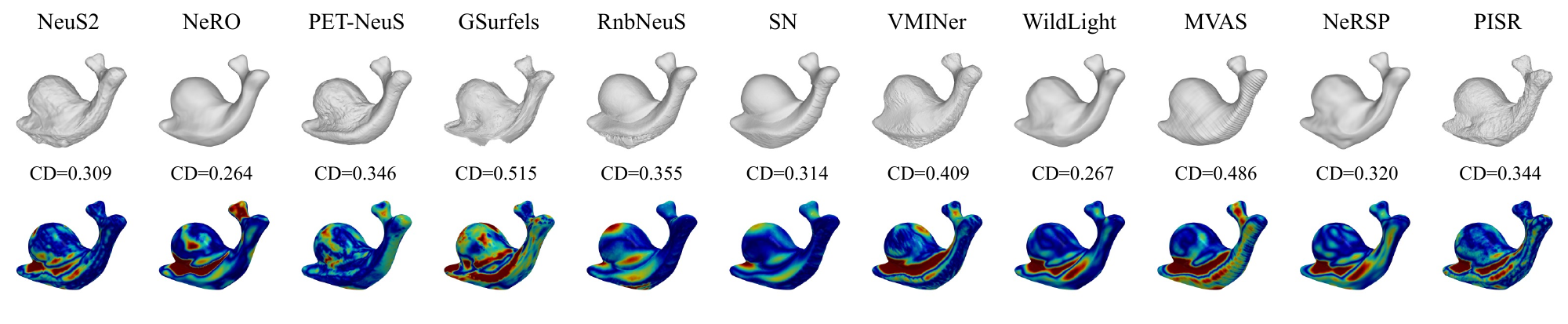}  
    \caption{Recovered shapes and error maps of {\sc Snail}.}
    \label{fig:snail_result}
\end{figure*}

\begin{figure*}
    \centering 
    \includegraphics[width=\linewidth]{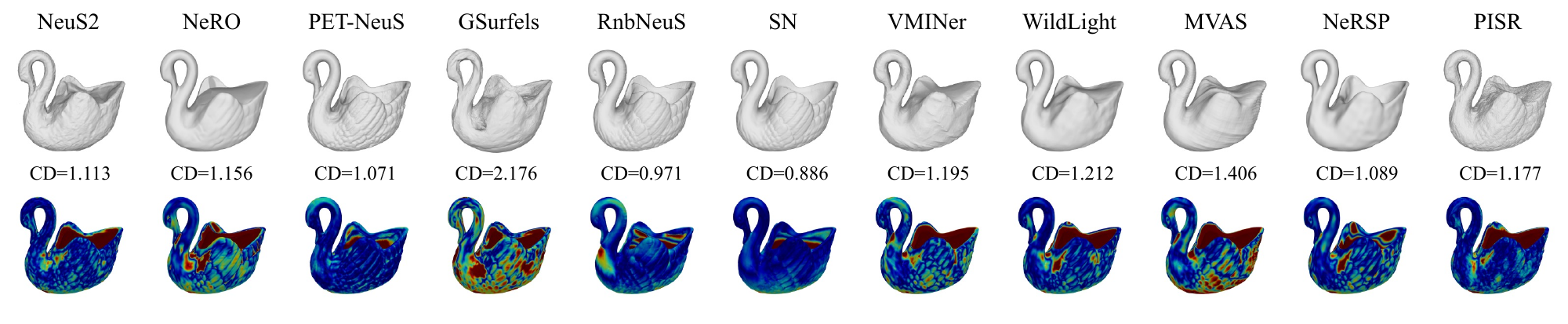}  
    \caption{Recovered shapes and error maps of {\sc Swan}.}
    \label{fig:swan_result}
\end{figure*}

\begin{figure*}
    \centering 
    \includegraphics[width=\linewidth]{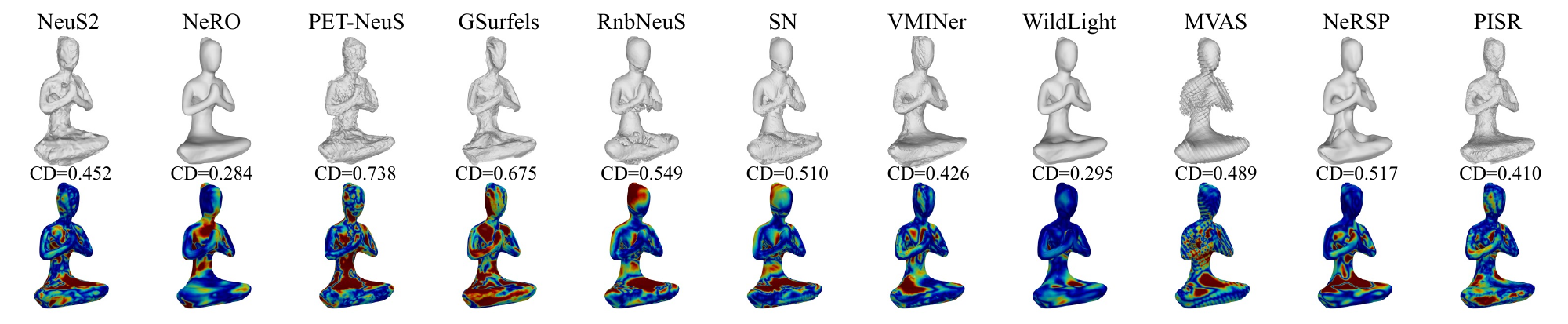}  
    \caption{Recovered shapes and error maps of {\sc Yoga}.}
    \label{fig:yoga_result}
\end{figure*}

\begin{figure*}
    \centering 
    \includegraphics[width=\linewidth]{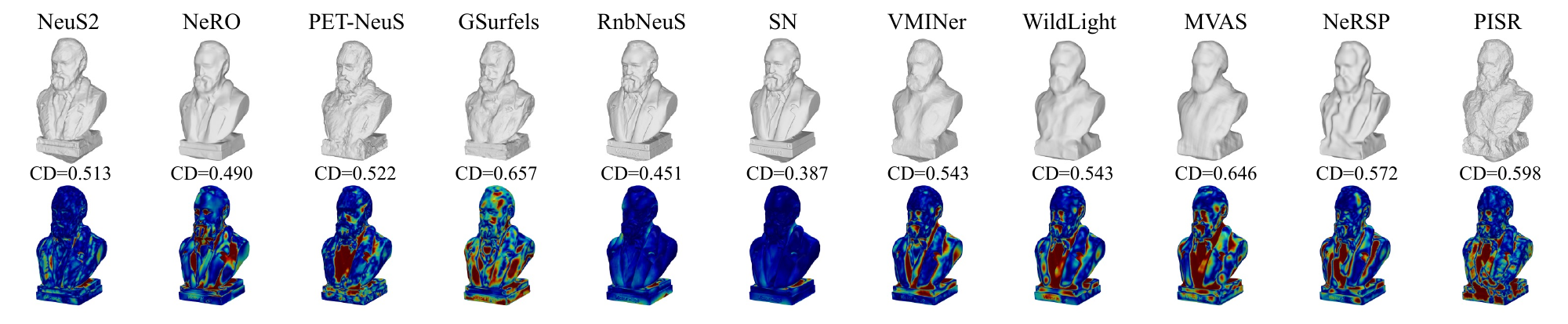}  
    \caption{Recovered shapes and error maps of {\sc Hugo}.}
    \label{fig:hugo_result}
\end{figure*}

\begin{figure*}
    \centering 
    \includegraphics[width=\linewidth]{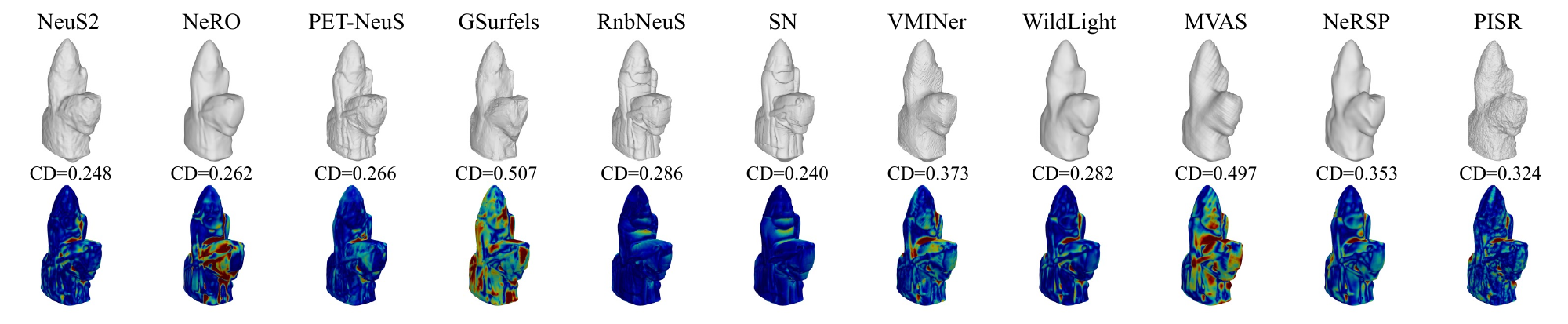}  
    \caption{Recovered shapes and error maps of {\sc Chess}.}
    \label{fig:chess_result}
\end{figure*}

\begin{figure*}
    \centering 
    \includegraphics[width=\linewidth]{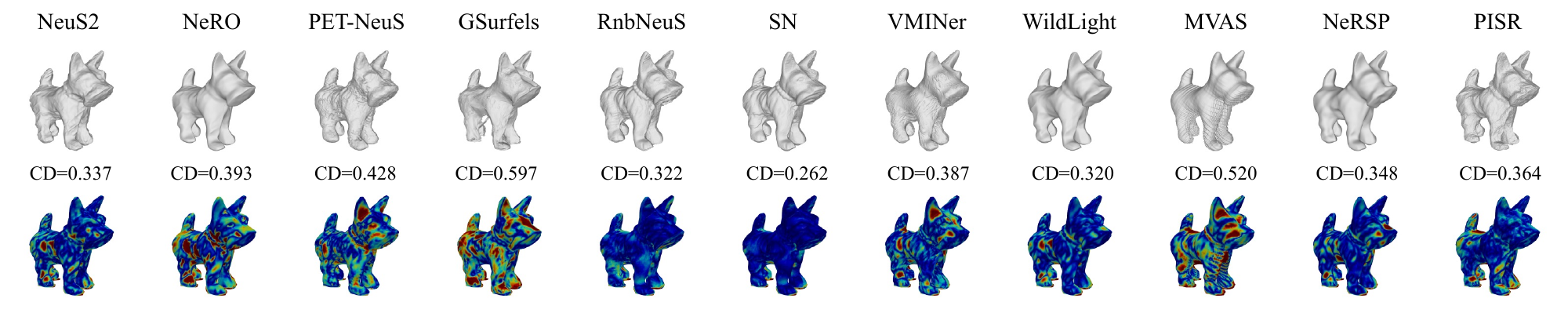}  
    \caption{Recovered shapes and error maps of {\sc BlackDog}.}
    \label{fig:blackdog_result}
\end{figure*}

\begin{figure*}
    \centering 
    \includegraphics[width=\linewidth]{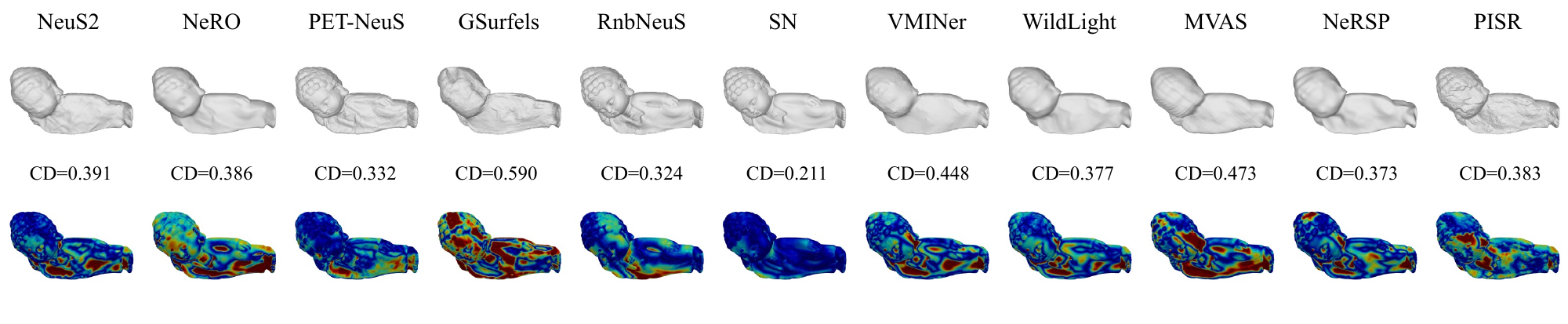}  
    \caption{Recovered shapes and error maps of {\sc Monk2}.}
    \label{fig:monk2_result}
\end{figure*}

\begin{figure*}
    \centering 
    \includegraphics[width=\linewidth]{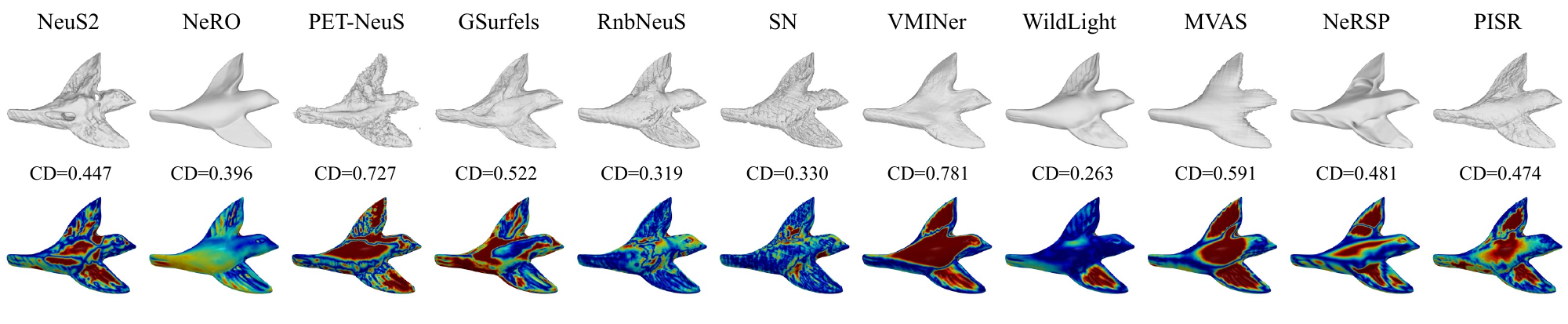}  
    \caption{Recovered shapes and error maps of {\sc Bird}.}
    \label{fig:bird_result}
\end{figure*}

\begin{figure*}
    \centering 
    \includegraphics[width=\linewidth]{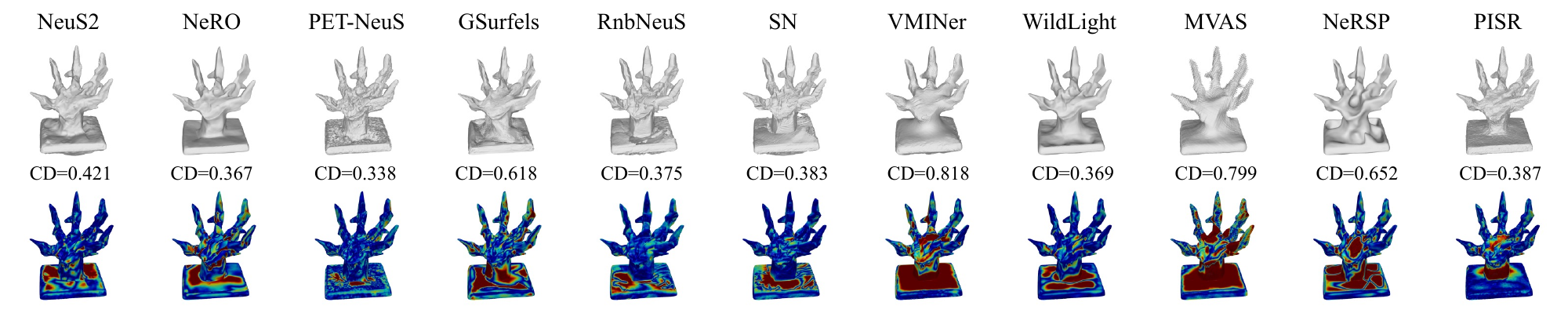}  
    \caption{Recovered shapes and error maps of {\sc Caw}.}
    \label{fig:caw_result}
\end{figure*}

\begin{figure*}
    \centering 
    \includegraphics[width=\linewidth]{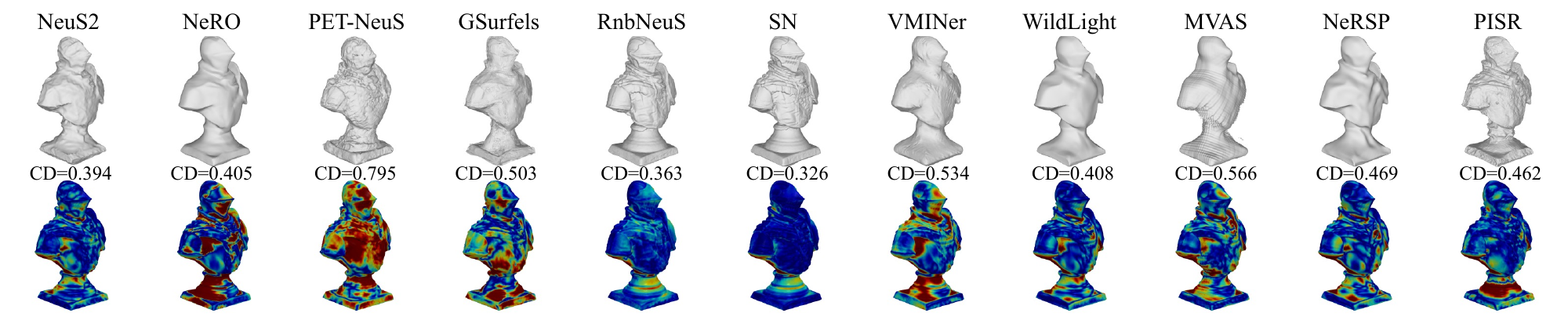}  
    \caption{Recovered shapes and error maps of {\sc Knight}.}
    \label{fig:knight_result}
\end{figure*}

\begin{figure*}
    \centering 
    \includegraphics[width=\linewidth]{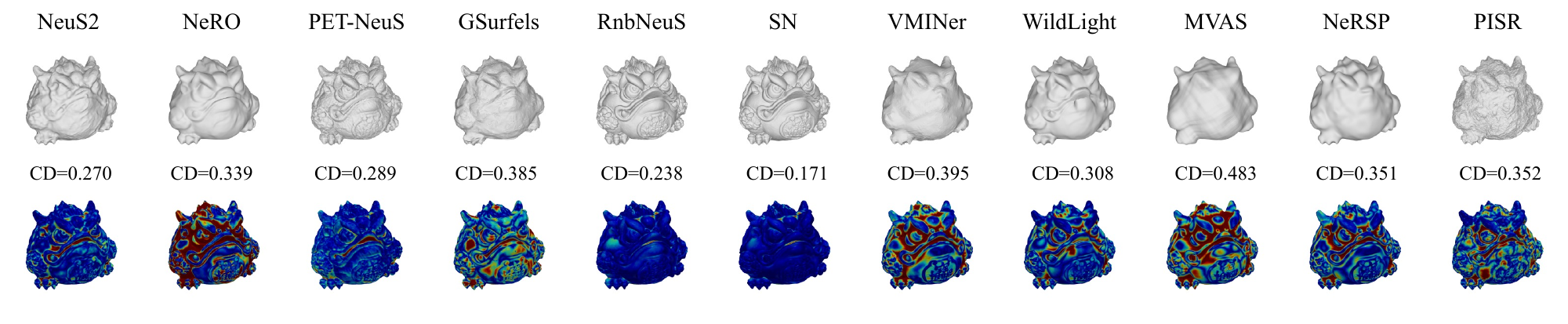}  
    \caption{Recovered shapes and error maps of {\sc Frog}.}
    \label{fig:frog_result}
\end{figure*}

\begin{figure*}
    \centering 
    \includegraphics[width=\linewidth]{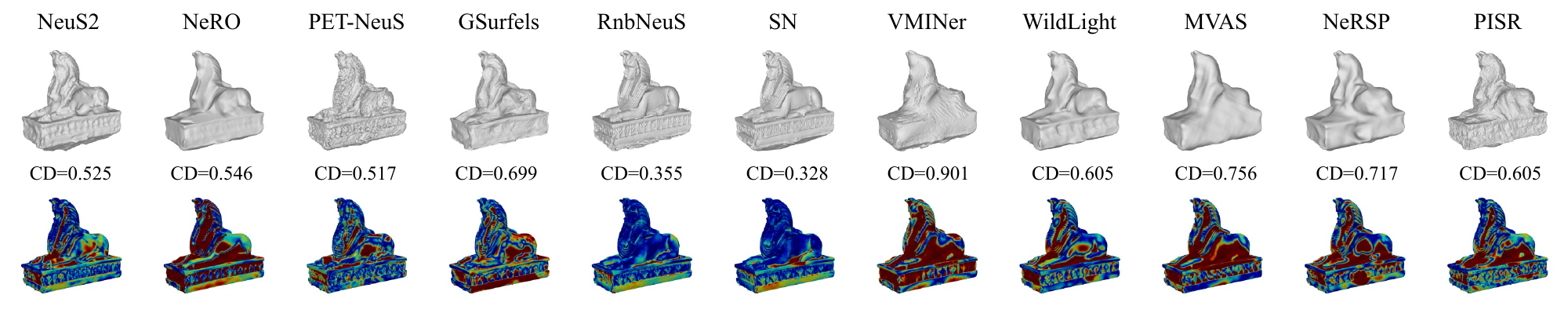}  
    \caption{Recovered shapes and error maps of {\sc Sphinx}.}
    \label{fig:sphinx_result}
\end{figure*}

\begin{figure*}
    \centering 
    \includegraphics[width=\linewidth]{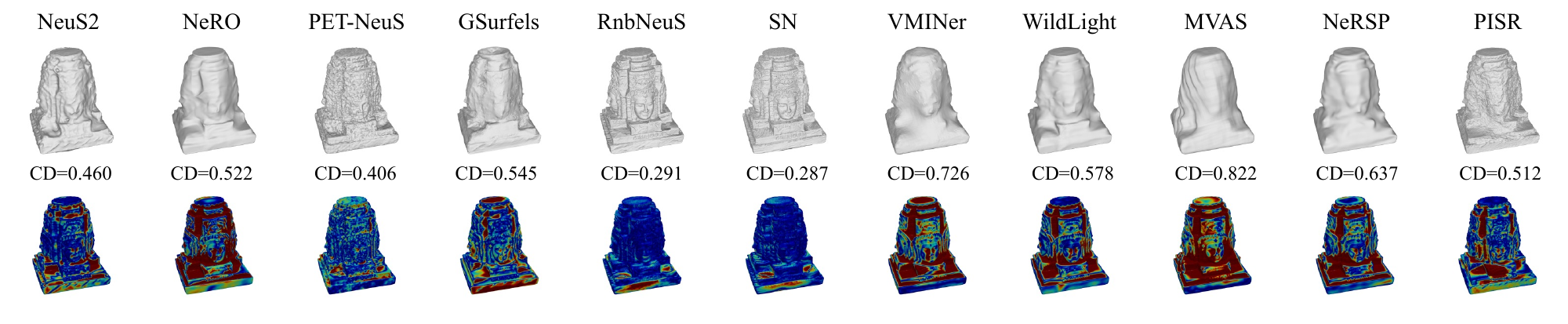}  
    \caption{Recovered shapes and error maps of {\sc Cambodia}.}
    \label{fig:cambodia_result}
\end{figure*}

\begin{figure*}
    \centering 
    \includegraphics[width=\linewidth]{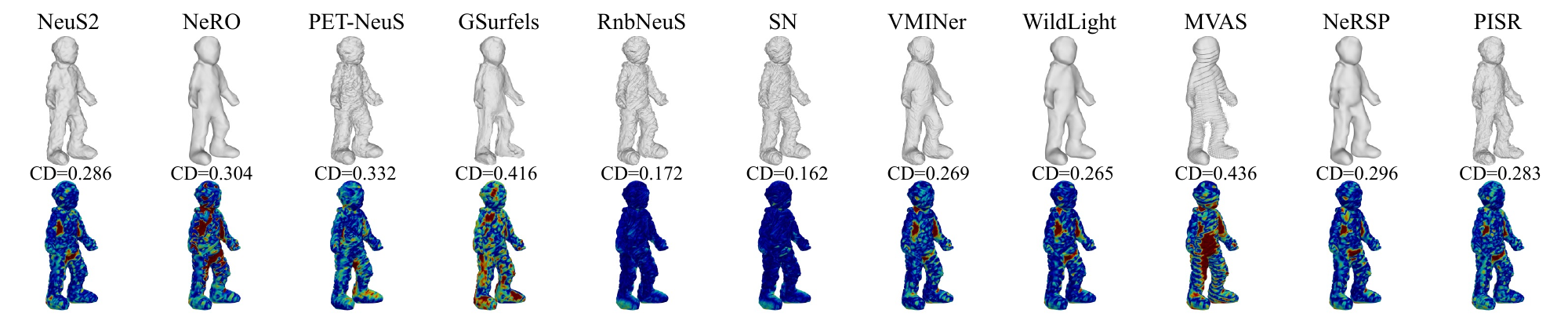}  
    \caption{Recovered shapes and error maps of {\sc Mummy}.}
    \label{fig:mummy_result}
\end{figure*}

\begin{figure*}
    \centering 
    \includegraphics[width=\linewidth]{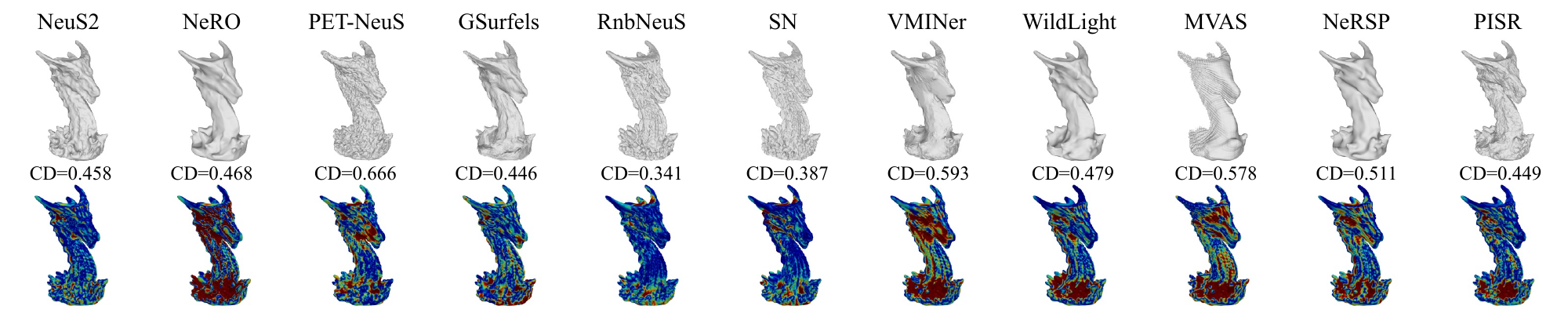}  
    \caption{Recovered shapes and error maps of {\sc Dragon}. }
    \label{fig:dragon_result}
\end{figure*}

From \fref{fig:face_result} to \fref{fig:dragon_result}, we present a comprehensive benchmark evaluation of the 11 methods reported in the main paper using \ours. Each method reconstructs 25 objects from 20 input views, and we provide the resulting 3D shapes along with their corresponding error maps. For better visualization, the maximum distance in the error maps is truncated at 1,mm.

{
    \small
    \bibliographystyle{tips/ieeenat_fullname}
    \bibliography{main}

@string{CVPR = "{Proc. of IEEE Conference on Computer Vision and Pattern Recognition (CVPR)}"}

@string{ECCV = "{Proc. of European Conference on Computer Vision (ECCV)}"}

@string{ICCV = "{Proc. of International Conference on Computer Vision (ICCV)}"}

@string{NIPS = "{Proc. of Annual Conference on Neural Information Processing Systems (NeurIPS)}"}

@string{I3DV = "{International Conference on 3D Vision (3DV)}"}

@string{SIGGRAPH = "{Proc. of SIGGRAPH}"}

@string{ToG = "{ACM Trans. on Graph.}"}

@article{neus,
	title={NeuS: Learning Neural Implicit Surfaces by Volume Rendering for Multi-view Reconstruction}, 
	author={Peng Wang and Lingjie Liu and Yuan Liu and Christian Theobalt and Taku Komura and Wenping Wang},
	journal=NIPS,
	year={2021}
}

@inproceedings{colmap,
	author={Sch\"{o}nberger, Johannes Lutz and Frahm, Jan-Michael},
	title={Structure-from-Motion Revisited},
	booktitle=CVPR,
	year={2016},
}

@inproceedings{pandora,
  title={Pandora: Polarization-aided neural decomposition of radiance},
  author={Dave, Akshat and Zhao, Yongyi and Veeraraghavan, Ashok},
  booktitle=ECCV,
  pages={538--556},
  year={2022},
  organization={Springer}
}

@inproceedings{nersp,
  title={NeRSP: Neural 3D Reconstruction for Reflective Objects with Sparse Polarized Images},
  author={Han, Yufei and Guo, Heng and Fukai, Koki and Santo, Hiroaki and Shi, Boxin and Okura, Fumio and Ma, Zhanyu and Jia, Yunpeng},
  booktitle=CVPR,
  pages={11821--11830},
  year={2024}
}

@inproceedings{neisf,
  title={NeISF: Neural Incident Stokes Field for Geometry and Material Estimation},
  author={Li, Chenhao and Ono, Taishi and Uemori, Takeshi and Mihara, Hajime and Gatto, Alexander and Nagahara, Hajime and Moriuchi, Yusuke},
  booktitle=CVPR,
  pages={21434--21445},
  year={2024}
}

@inproceedings{mvas,
  title={Multi-view azimuth stereo via tangent space consistency},
  author={Cao, Xu and Santo, Hiroaki and Okura, Fumio and Matsushita, Yasuyuki},
  booktitle=CVPR,
  pages={825--834},
  year={2023}
}

@inproceedings{psnerf,
  title={Ps-nerf: Neural inverse rendering for multi-view photometric stereo},
  author={Yang, Wenqi and Chen, Guanying and Chen, Chaofeng and Chen, Zhenfang and Wong, Kwan-Yee K},
  booktitle=ECCV,
  pages={266--284},
  year={2022},
  organization={Springer}
}

@inproceedings{rnbneus,
  title={RNb-NeuS: Reflectance and Normal-based Multi-View 3D Reconstruction},
  author={Brument, Baptiste and Bruneau, Robin and Qu{\'e}au, Yvain and M{\'e}lou, Jean and Lauze, Fran{\c{c}}ois Bernard and Durou, Jean-Denis and Calvet, Lilian},
  booktitle=CVPR,
  pages={5230--5239},
  year={2024}
}

@inproceedings{supernormal,
  title={SuperNormal: Neural Surface Reconstruction via Multi-View Normal Integration},
  author={Cao, Xu and Taketomi, Takafumi},
  booktitle=CVPR,
  pages={20581--20590},
  year={2024}
}

@inproceedings{Mvpsnet,
  title={Mvpsnet: Fast generalizable multi-view photometric stereo},
  author={Zhao, Dongxu and Lichy, Daniel and Perrin, Pierre-Nicolas and Frahm, Jan-Michael and Sengupta, Soumyadip},
  booktitle=ICCV,
  pages={12525--12536},
  year={2023}
}

@inproceedings{uamvps,
  title={Uncertainty-aware deep multi-view photometric stereo},
  author={Kaya, Berk and Kumar, Suryansh and Oliveira, Carlos and Ferrari, Vittorio and Van Gool, Luc},
  booktitle={Proceedings of the IEEE/CVF conference on computer vision and pattern recognition},
  pages={12601--12611},
  year={2022}
}

@inproceedings{vminer,
  title={VMINer: Versatile Multi-view Inverse Rendering with Near-and Far-field Light Sources},
  author={Fei, Fan and Tang, Jiajun and Tan, Ping and Shi, Boxin},
  booktitle=CVPR,
  pages={11800--11809},
  year={2024}
}

@inproceedings{wildlight,
  title={Wildlight: In-the-wild inverse rendering with a flashlight},
  author={Cheng, Ziang and Li, Junxuan and Li, Hongdong},
  booktitle=CVPR,
  pages={4305--4314},
  year={2023}
}

@misc{bmvps,
      title={Multi-View Photometric Stereo: A Robust Solution and Benchmark Dataset for Spatially Varying Isotropic Materials}, 
      author={Min Li and Zhenglong Zhou and Zhe Wu and Boxin Shi and Changyu Diao and Ping Tan},
      year={2020},
      eprint={2001.06659},
      archivePrefix={arXiv},
      primaryClass={cs.CV},
      url={https://arxiv.org/abs/2001.06659}, 
}

@article{nero,
  title={Nero: Neural geometry and brdf reconstruction of reflective objects from multiview images},
  author={Liu, Yuan and Wang, Peng and Lin, Cheng and Long, Xiaoxiao and Wang, Jiepeng and Liu, Lingjie and Komura, Taku and Wang, Wenping},
  journal=ToG,
  volume={42},
  number={4},
  pages={1--22},
  year={2023},
  publisher={ACM New York, NY, USA}
}

@article{relightable3d,
  title={Relightable 3d gaussian: Real-time point cloud relighting with brdf decomposition and ray tracing},
  author={Gao, Jian and Gu, Chun and Lin, Youtian and Zhu, Hao and Cao, Xun and Zhang, Li and Yao, Yao},
  journal={arXiv preprint arXiv:2311.16043},
  year={2023}
}

@inproceedings{neus2,
  title={Neus2: Fast learning of neural implicit surfaces for multi-view reconstruction},
  author={Wang, Yiming and Han, Qin and Habermann, Marc and Daniilidis, Kostas and Theobalt, Christian and Liu, Lingjie},
  booktitle=CVPR,
  pages={3295--3306},
  year={2023}
}

@inproceedings{neuralangelo,
  title={Neuralangelo: High-fidelity neural surface reconstruction},
  author={Li, Zhaoshuo and M{\"u}ller, Thomas and Evans, Alex and Taylor, Russell H and Unberath, Mathias and Liu, Ming-Yu and Lin, Chen-Hsuan},
  booktitle=CVPR,
  pages={8456--8465},
  year={2023}
}

@inproceedings{petneus,
  title={Pet-neus: Positional encoding tri-planes for neural surfaces},
  author={Wang, Yiqun and Skorokhodov, Ivan and Wonka, Peter},
  booktitle=CVPR,
  pages={12598--12607},
  year={2023}
}

@inproceedings{svolsdf,
  title={S-volsdf: Sparse multi-view stereo regularization of neural implicit surfaces},
  author={Wu, Haoyu and Graikos, Alexandros and Samaras, Dimitris},
  booktitle=ICCV,
  pages={3556--3568},
  year={2023}
}

@inproceedings{sparseneus,
  title={Sparseneus: Fast generalizable neural surface reconstruction from sparse views},
  author={Long, Xiaoxiao and Lin, Cheng and Wang, Peng and Komura, Taku and Wang, Wenping},
  booktitle=ECCV,
  pages={210--227},
  year={2022},
  organization={Springer}
}

@inproceedings{dtumvs,
  title={Large scale multi-view stereopsis evaluation},
  author={Jensen, Rasmus and Dahl, Anders and Vogiatzis, George and Tola, Engin and Aan{\ae}s, Henrik},
  booktitle=CVPR,
  pages={406--413},
  year={2014}
}

@article{stanfordorb,
  title={Stanford-ORB: a real-world 3d object inverse rendering benchmark},
  author={Kuang, Zhengfei and Zhang, Yunzhi and Yu, Hong-Xing and Agarwala, Samir and Wu, Elliott and Wu, Jiajun and others},
  journal=NIPS,
  volume={36},
  year={2024}
}

@inproceedings{rene,
  title={Relight my nerf: A dataset for novel view synthesis and relighting of real world objects},
  author={Toschi, Marco and De Matteo, Riccardo and Spezialetti, Riccardo and De Gregorio, Daniele and Di Stefano, Luigi and Salti, Samuele},
  booktitle=CVPR,
  pages={20762--20772},
  year={2023}
}

@article{openillumination,
  title={Openillumination: A multi-illumination dataset for inverse rendering evaluation on real objects},
  author={Liu, Isabella and Chen, Linghao and Fu, Ziyang and Wu, Liwen and Jin, Haian and Li, Zhong and Wong, Chin Ming Ryan and Xu, Yi and Ramamoorthi, Ravi and Xu, Zexiang and others},
  journal=NIPS,
  volume={36},
  year={2024}
}

@inproceedings{old,
  title={Objects With Lighting: A Real-World Dataset for Evaluating Reconstruction and Rendering for Object Relighting},
  author={Ummenhofer, Benjamin and Agrawal, Sanskar and Sepulveda, Rene and Lao, Yixing and Zhang, Kai and Cheng, Tianhang and Richter, Stephan and Wang, Shenlong and Ros, German},
  booktitle=I3DV,
  pages={137--147},
  year={2024},
  organization={IEEE}
}

@inproceedings{mvimgnet,
	title={Mvimgnet: A large-scale dataset of multi-view images},
	author={Yu, Xianggang and Xu, Mutian and Zhang, Yidan and Liu, Haolin and Ye, Chongjie and Wu, Yushuang and Yan, Zizheng and Zhu, Chenming and Xiong, Zhangyang and Liang, Tianyou and others},
	booktitle=CVPR,
	pages={9150--9161},
	year={2023}
}

@inproceedings{mobilebrick,
	title={Mobilebrick: Building lego for 3d reconstruction on mobile devices},
	author={Li, Kejie and Bian, Jia-Wang and Castle, Robert and Torr, Philip HS and Prisacariu, Victor Adrian},
	booktitle=CVPR,
	pages={4892--4901},
	year={2023}
}

@inproceedings{co3d,
    title     = {Common Objects in 3D: Large-Scale Learning and Evaluation of Real-Life 3D Category Reconstruction},
    author    = {Reizenstein, Jeremy and Shapovalov, Roman and Henzler, Philipp and Sbordone, Luca and Labatut, Patrick and Novotny, David},
    booktitle =ICCV,
    pages     = {10901--10911},
    year      = {2021}
}

@article{nerf,
	title={Nerf: Representing scenes as neural radiance fields for view synthesis},
	author={Mildenhall, Ben and Srinivasan, Pratul P and Tancik, Matthew and Barron, Jonathan T and Ramamoorthi, Ravi and Ng, Ren},
	journal={Communications of the ACM},
	volume={65},
	number={1},
	pages={99--106},
	year={2021},
	publisher={ACM New York, NY, USA}
}

@article{volsdf,
	title={Volume rendering of neural implicit surfaces},
	author={Yariv, Lior and Gu, Jiatao and Kasten, Yoni and Lipman, Yaron},
	journal=NIPS,
	volume={34},
	pages={4805--4815},
	year={2021}
}

@inproceedings{refneus,
	title={Ref-neus: Ambiguity-reduced neural implicit surface learning for multi-view reconstruction with reflection},
	author={Ge, Wenhang and Hu, Tao and Zhao, Haoyu and Liu, Shu and Chen, Ying-Cong},
	booktitle=CVPR,
	pages={4251--4260},
	year={2023}
}

@inproceedings{refnerf,
	title={Ref-nerf: Structured view-dependent appearance for neural radiance fields},
	author={Verbin, Dor and Hedman, Peter and Mildenhall, Ben and Zickler, Todd and Barron, Jonathan T and Srinivasan, Pratul P},
	booktitle=CVPR,
	pages={5481--5490},
	year={2022},
	organization={IEEE}
}

@Article{3Dgaussians,
	author       = {Kerbl, Bernhard and Kopanas, Georgios and Leimk{\"u}hler, Thomas and Drettakis, George},
	title        = {3D Gaussian Splatting for Real-Time Radiance Field Rendering},
	journal      = TOG,
	number       = {4},
	volume       = {42},
	month        = {July},
	year         = {2023},
	url          = {https://repo-sam.inria.fr/fungraph/3d-gaussian-splatting/}
}

@inproceedings{gaussianS,
	title={High-quality surface reconstruction using gaussian surfels},
	author={Dai, Pinxuan and Xu, Jiamin and Xie, Wenxiang and Liu, Xinguo and Wang, Huamin and Xu, Weiwei},
	booktitle=SIGGRAPH,
	pages={1--11},
	year={2024}
}

@inproceedings{sdm,
	title={Scalable, detailed and mask-free universal photometric stereo},
	author={Ikehata, Satoshi},
	booktitle=CVPR,
	pages={13198--13207},
	year={2023}
}

@article{hardy2024uni,
  title={Uni MS-PS: A multi-scale encoder-decoder transformer for universal photometric stereo},
  author={Hardy, Cl{\'e}ment and Qu{\'e}au, Yvain and Tschumperl{\'e}, David},
  journal={Computer Vision and Image Understanding},
  volume={248},
  pages={104093},
  year={2024},
  publisher={Elsevier}
}

@article{pisr,
	title={PISR: Polarimetric Neural Implicit Surface Reconstruction for Textureless and Specular Objects},
	author={Chen, Guangcheng and He, Yicheng and He, Li and Zhang, Hong},
	journal=ECCV,
	year={2024}
}

@article{ravi2020pytorch3d,
	author = {Nikhila Ravi and Jeremy Reizenstein and David Novotny and Taylor Gordon
	and Wan-Yen Lo and Justin Johnson and Georgia Gkioxari},
	title = {Accelerating 3D Deep Learning with PyTorch3D},
	journal = {arXiv:2007.08501},
	year = {2020},
}

@inproceedings{kirillov2023segment,
	title={Segment anything},
	author={Kirillov, Alexander and Mintun, Eric and Ravi, Nikhila and Mao, Hanzi and Rolland, Chloe and Gustafson, Laura and Xiao, Tete and Whitehead, Spencer and Berg, Alexander C and Lo, Wan-Yen and others},
	booktitle=CVPR,
	pages={4015--4026},
	year={2023}
}

@inproceedings{voynov2023multi,
  title={Multi-sensor large-scale dataset for multi-view 3D reconstruction},
  author={Voynov, Oleg and Bobrovskikh, Gleb and Karpyshev, Pavel and Galochkin, Saveliy and Ardelean, Andrei-Timotei and Bozhenko, Arseniy and Karmanova, Ekaterina and Kopanev, Pavel and Labutin-Rymsho, Yaroslav and Rakhimov, Ruslan and others},
  booktitle={Proceedings of the IEEE/CVF Conference on Computer Vision and Pattern Recognition},
  pages={21392--21403},
  year={2023}
}

@inproceedings{wu2023omniobject3d,
  title={Omniobject3d: Large-vocabulary 3d object dataset for realistic perception, reconstruction and generation},
  author={Wu, Tong and Zhang, Jiarui and Fu, Xiao and Wang, Yuxin and Ren, Jiawei and Pan, Liang and Wu, Wayne and Yang, Lei and Wang, Jiaqi and Qian, Chen and others},
  booktitle={Proceedings of the IEEE/CVF Conference on Computer Vision and Pattern Recognition},
  pages={803--814},
  year={2023}
}

@inproceedings{besl1992method,
  title={Method for registration of 3-D shapes},
  author={Besl, Paul J and McKay, Neil D},
  booktitle={Sensor fusion IV: control paradigms and data structures},
  volume={1611},
  pages={586--606},
  year={1992},
  organization={Spie}
}

@inproceedings{heep2024adaptive,
  title={An Adaptive Screen-Space Meshing Approach for Normal Integration},
  author={Heep, Moritz and Zell, Eduard},
  booktitle={European Conference on Computer Vision},
  pages={445--461},
  year={2024},
  organization={Springer}
}

@manual{blender,
  title        = {Blender - a 3D modelling and rendering package},
  author       = {{Blender Online Community}},
  organization = {Blender Foundation},
  address      = {Amsterdam, The Netherlands},
  year         = {2025},
  url          = {https://www.blender.org}
}

@inproceedings{cao2021normal,
  title={Normal Integration via Inverse Plane Fitting With Minimum Point-to-Plane Distance},
  author={Cao, Xu and Shi, Boxin and Okura, Fumio and Matsushita, Yasuyuki},
  booktitle={Proceedings of the IEEE/CVF Conference on Computer Vision and Pattern Recognition},
  pages={2382--2391},
  year={2021}
}

@inproceedings{liu2023neudf,
  title={Neudf: Leaning neural unsigned distance fields with volume rendering},
  author={Liu, Yu-Tao and Wang, Li and Yang, Jie and Chen, Weikai and Meng, Xiaoxu and Yang, Bo and Gao, Lin},
  booktitle={Proceedings of the IEEE/CVF conference on computer vision and pattern recognition},
  pages={237--247},
  year={2023}
}
}

% \input{sec/X_suppl}

% WARNING: do not forget to delete the supplementary pages from your submission 
% \input{sec/X_suppl}

\end{document}